\documentclass[final,3p,times,authoryear]{elsarticle}

\usepackage{amssymb}
\usepackage{amsmath}
\usepackage{microtype}
\usepackage{graphicx}
\usepackage{subfigure}
\usepackage{booktabs}
\usepackage{subcaption}

\usepackage{algorithm}
\usepackage{algpseudocode}
\usepackage{multicol}
\usepackage{multirow}
\usepackage{hyperref}
\hypersetup{
colorlinks=false,
linkcolor=black,
anchorcolor=black,
citecolor=black
}

\journal{Neural Networks}

\begin{document}

\begin{frontmatter}



\title{Repetitive Contrastive Learning Enhances Mamba's Selectivity in Time Series Prediction} 

\cortext[cor1]{Corresponding author}
\fntext[equal]{Same Contribution}
\author[equal,11,22]{Wenbo Yan}
\author[equal,11]{Hanzhong Cao}
\author[11,33,44]{Ying Tan\corref{cor1}}

\affiliation[11]{organization={School of Intelligence Science and Technology, Peking University, Beijing}
}

\affiliation[22]{organization={Computational Intelligence Laboratory}
}
\affiliation[33]{organization={Institute for Artificial Intelligence}
}
\affiliation[44]{organization={State Key Laboratory of General Artificial Intelligence, Peking University, Beijing}
}

\begin{abstract}
Long sequence prediction is a key challenge in time series forecasting. While Mamba-based models have shown strong performance due to their sequence selection capabilities, they still struggle with insufficient focus on critical time steps and incomplete noise suppression, caused by limited selective abilities. To address this, we introduce Repetitive Contrastive Learning (RCL), a token-level contrastive pretraining framework aimed at enhancing Mamba's selective capabilities. RCL pretrains a single Mamba block to strengthen its selective abilities and then transfers these pretrained parameters to initialize Mamba blocks in various backbone models, improving their temporal prediction performance. RCL uses sequence augmentation with Gaussian noise and applies inter-sequence and intra-sequence contrastive learning to help the Mamba module prioritize information-rich time steps while ignoring noisy ones. Extensive experiments show that RCL consistently boosts the performance of backbone models, surpassing existing methods and achieving state-of-the-art results. Additionally, we propose two metrics to quantify Mamba's selective capabilities, providing theoretical, qualitative, and quantitative evidence for the improvements brought by RCL.

\end{abstract}








\begin{keyword}


Time Series Forecasting \sep Long Sequence Prediction \sep Mamba \sep Repetitive Contrastive Learning  \sep  Selectivity Measurement 
\end{keyword}

\end{frontmatter}



\section{Introduction}
\label{sec1}

Time series forecasting (TSF) has become indispensable across critical domains such as financial markets \citep{li2023mastermarketguidedstocktransformer}, traffic management \citep{cheng2023deeptransportlearningspatialtemporaldependency}, electricity consumption \citep{Sun2023}, scientific computing \citep{10.1145/3682060}, and weather prediction \citep{ZHANG2022111768}. TSF leverages sequential data—often irregular, incomplete, or noisy—to predict future trends based on past observations. Deep learning has advanced the field substantially, with increasing attention on architectural innovations, particularly transformer-based models \citep{wen2023transformerstimeseriessurvey}. Yet, conventional models like CNNs and MLPs remain competitive due to their robustness against noise and ability to capture local patterns \citep{zeng2022transformerseffectivetimeseries}. More recently, the Mamba model \citep{gu2024mambalineartimesequencemodeling}, powered by selective state-space mechanisms \citep{huang2024mlmambaefficientmultimodallarge,li2024mambandselectivestatespace}, has gained traction, inspiring variants such as TimeMachine \citep{ahamed2024timemachinetimeseriesworth} and Bi-Mamba \citep{liang2024bimambabidirectionalmambatime}.

Despite these advances, fully reliable forecasting remains elusive. The challenges stem from the opaque generative processes of time series data, exacerbated by uneven sampling, missing or redundant entries, and unpredictable disturbances \citep{zhu2023networkedtimeseriesprediction,ramponi2019tcganconditionalgenerativeadversarial}. Transformer architectures, though powerful in NLP, often underperform in these settings, struggling with sequential irregularity and noise. While state-space approaches like Mamba alleviate complexity and preserve long-range dependencies, they often overlook a core challenge: enabling models to identify and prioritize the most informative timesteps. Addressing this issue is vital for robust and interpretable forecasting \citep{nam2024adversariallearningapproachirregular}. In particular, as we will discuss next, Mamba’s selective mechanism—though promising—introduces its own limitations when applied to time series forecasting.

\begin{figure}
    \centering
    \includegraphics[width=\linewidth]{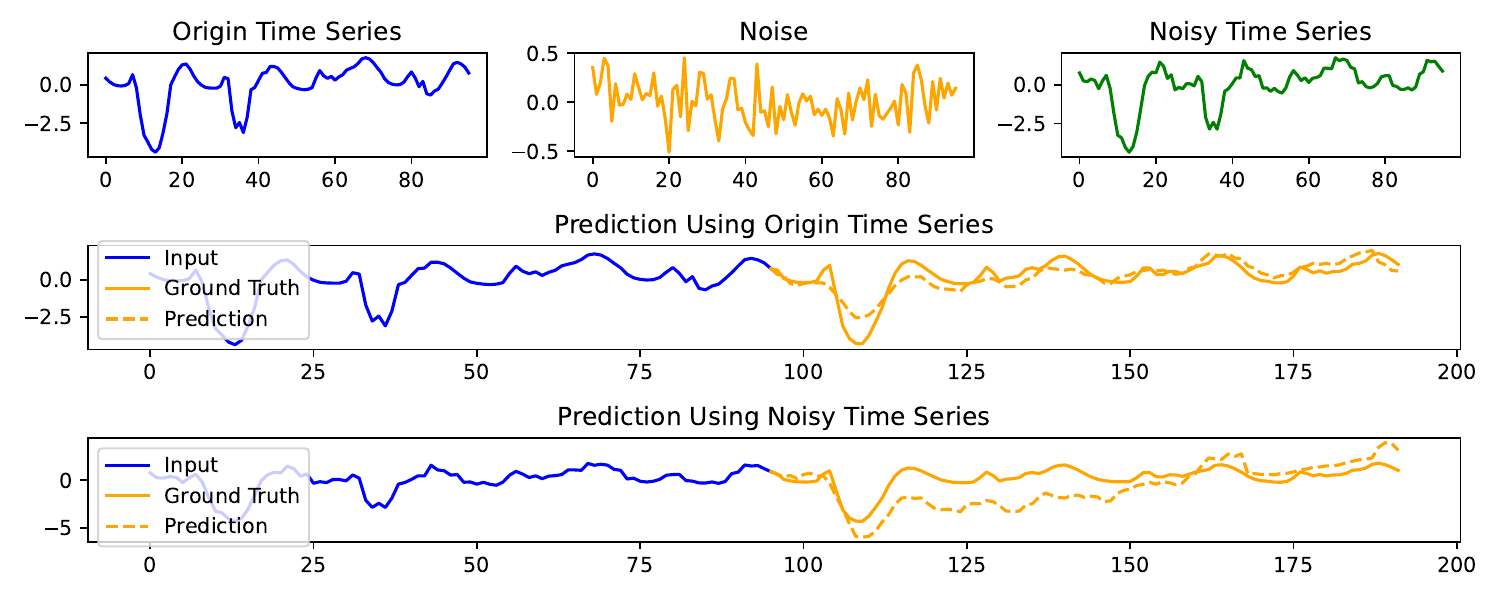}
    \caption{Impact of Noise Sensitivity on Prediction Results}
    \label{fig:motivation}
\end{figure}
Although Mamba introduces a degree of selectivity by generating its state transition and input matrices dynamically based on time steps, this very mechanism also makes it more sensitive to temporal fluctuations and noise. 
As illustrated in Figure \ref{fig:motivation}, Mamba exhibits two key limitations when predicting time series. On one hand, it fails to effectively focus on salient time steps, resulting in an inability to fit extreme cases such as sharp declines or rapid rises. On the other hand, Mamba is highly sensitive to noise, where even minor noise disturbances can lead to significant deviations in prediction results. This is attributed to insufficient selective capabilities of Mamba, causing noise to accumulate progressively during sequence modeling, ultimately amplifying small noise into large deviations. These limitations arise from its origins in natural language processing (NLP), where each token typically carries rich semantic meaning. In contrast, time series data often exhibit irregular sampling, low signal-to-noise ratios, and a lack of contextual semantics, making them fundamentally different from NLP data. Prior studies have also shown that directly applying NLP-inspired architectures to time series tasks yields suboptimal results \citep{zhang2024integrationmambatransformer}.

To address these shortcomings, we propose a token-level pre-training method specifically designed to enhance the initialization of the Mamba block, termed Repetitive Contrastive Learning (RCL). This approach aims to endow the Mamba architecture with stronger selective capabilities—enabling it to better attend to salient time steps while ignoring irrelevant or noisy ones. Importantly, the resulting initialization parameters are architecture-agnostic and can be flexibly applied to any model employing the Mamba block, thereby strengthening its capacity to model complex temporal dependencies.


Specifically, RCL is a novel pre-training framework comprising two key steps: Repeating Sequence Augmentation and Repetitive Contrastive Learning, meticulously designed to address the dual challenges of denoising and memorization. In the Repeating Sequence Augmentation step, each token in the sequence is duplicated and perturbed with Gaussian noise to simulate the irregular and redundant patterns commonly observed in real-world time series. Subsequently, during the Repetitive Contrastive Learning phase, we enhance the model’s ability to ignore noisy timesteps through \textbf{intra-sequence contrast}, thereby suppressing spurious fluctuations. Simultaneously, \textbf{inter-sequence contrast} ensures the consistency of temporal features across sequences of varying lengths, preventing the loss of temporal variation extraction capabilities caused by repetitive augmentation and noise introduction. These two contrastive mechanisms collectively imbue Mamba with sharper selective capabilities and higher temporal fidelity, ultimately enabling more robust performance across diverse time series forecasting tasks.

We integrate this training paradigm into the training process of Mamba-based models by employing RCL to train a single Mamba block, obtaining parameters with enhanced selectivity. These parameters are then used as the initialization parameters for all Mamba blocks within the model, thereby boosting the overall predictive capability. Experimental evaluations on multiple Mamba-based models demonstrate that our approach significantly enhances the predictive performance of the backbone models, achieving state-of-the-art (SOTA) results without incurring additional memory overhead. Furthermore, we explore module replacement and parameter freezing strategies to maximize transferability and training stability.

In summary, our main contributions are as follows: 

\begin{itemize}



\item We propose a token-level training paradigm called Repeating Sequence Augmentation. By repeating timesteps and introducing noise, combined with intra-sequence and inter-sequence contrastive learning, the parameters of the Mamba block acquire the ability to identify key timesteps and ignore noise.

\item We integrate RCL into the training process of various Mamba-based models, utilizing the parameters obtained from RCL as the initialization for the backbone model’s parameters, further enhancing the temporal prediction capabilities of the backbone model.


\item Experiments demonstrate that RCL can significantly improve the performance of backbone models, and its broad effectiveness for Mamba-based models is verified without incurring additional memory overhead. And the impact of different parameter replacement methods and freezing techniques is analyzed through experiments.




    
    
        
\end{itemize}

\section{Preliminary}

\subsection{Multivariate Time Series Forecasting}

Multivariate time series forecasting involves predicting future values of multiple interrelated time-dependent variables based on their historical data. Unlike univariate time series forecasting, which focuses on a single variable, multivariate forecasting accounts for interactions and correlations between multiple variables to improve prediction accuracy and insightfulness.

A multivariate time series forecasting problem can be formally represented with an input time series denoted as $\textbf{X} \in \mathbb{R}^{T_{\text{in}} \times F}$, where $T_{\text{in}}$ is the input sequence length (number of time steps) and $F$ represents the number of features or variables at each time step. The prediction target is represented as $\textbf{Y} \in \mathbb{R}^{T_{\text{out}} \times F}$, where $T_{\text{out}}$ denotes the output sequence length for which forecasts are made.



\subsection{Mamba Block}\label{sec:mambablock}
\begin{figure}[htbp]
    \centering
    \includegraphics[width=0.8\linewidth]{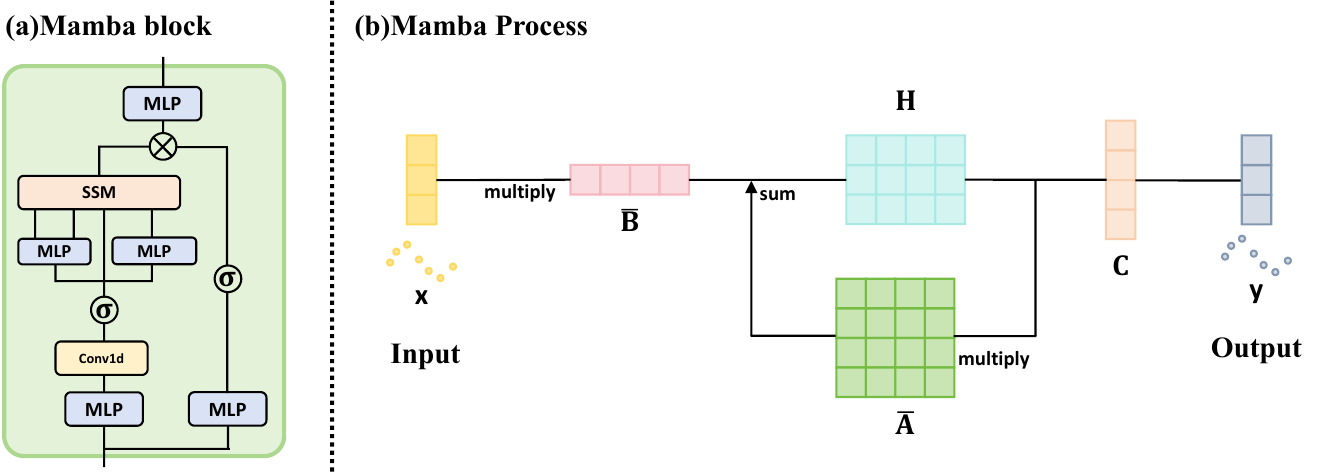}
    \caption{(a)The structure of the mamba block. (b)The Process of Sequence Modeling with Mamba}
    \label{fig:mambaarch}
\end{figure}

The Mamba block, \citep{gu2024mambalineartimesequencemodeling}, consists of two parts : selection and State Space Model (SSM), as shown in Fig. \ref{fig:mambaarch}(a). Firstly, the input $\textbf{X}$ undergoes a one-dimensional convolution (Conv1d) to extract local features, followed by Linear Projection that maps it to matrices $\textbf{B}$, $\textbf{C}$, and $\Delta$.

\begin{equation}
    \begin{aligned}
        \textbf{X}_c &= \text{fc}(\sigma(\text{Conv1d}(\textbf{X}))) \\
        \textbf{B} &= \text{fc}(\textbf{X}_c), \quad \textbf{C} = \text{fc}(\textbf{X}_c) \\
        \Delta &= \text{softplus}(\text{fc}(\textbf{X}_c) + \textbf{A})
    \end{aligned}
\end{equation}

where $\sigma$ is SiLU activation function and softplus means the Softplus activation functions, and $\textbf{A}$ is an optimizable matrix. Then, matrices $\textbf{A}$ and $\textbf{B}$ are discretized into $ \overline{\textbf{A}}$, $ \overline{\textbf{B}}$,
\begin{equation}
    \begin{aligned}
\overline{\textbf{A}} &= \exp {(\Delta\textbf{A})}\\
\overline{\textbf{B}} &= (\exp {(\Delta\textbf{A})}-\textbf{I})(\Delta\textbf{A})^{-1}(\Delta\textbf{B})
    \end{aligned}
\end{equation}

Finally, Mamba inputs $\overline{\textbf{A}}$, $\overline{\textbf{B}}$, $\textbf{C}$, $\Delta$, and $\textbf{X}$ into the SSM with residual connections and passes the output through a Linear Projection.

\begin{equation}
    \begin{aligned}
        \textbf{H} = \text{fc}(\text{SSM}(\overline{\textbf{A}}, \overline{\textbf{B}}, \textbf{X}) \cdot \sigma(\text{fc}(\textbf{X})))
    \end{aligned}
\end{equation}

where $\text{fc}$ is fully connected layers, and $\sigma$ is SiLU activation function. The process of sequence modeling with Mamba is illustrated in Fig. \ref{fig:mambaarch}(b). The computational process of the State Space Model (SSM) can be succinctly represented as follows:
\begin{equation}
    \begin{aligned}
        \mathbf{h}_t &= \mathbf{\overline{A}} \mathbf{h}_{t-1} + \mathbf{\overline{B}} \mathbf{x}_t \\ 
        \mathbf{o}_t &= \mathbf{C} \mathbf{h}_t
    \end{aligned}\label{con:ssm}
\end{equation}

\subsection{Definition of Mamba's Selectivity}\label{mamba_select}



The selectivity of Mamba primarily stems from its unique Selective State Space Model (SSM). The computational process of the SSM is shown in Eq.\ref{con:ssm}. Unlike traditional SSMs \citep{gu2022efficientlymodelinglongsequences}, the state transition matrix \(\overline{\textbf{A}}\) and the input matrix \(\overline{\textbf{B}}\) are derived from the current timestep. As a result, the state transition matrix and input matrix generated based on the timestep can selectively decide whether to retain more historical state information or incorporate more information from the current timestep \citep{gu2024mambalineartimesequencemodeling}. 

We uniformly define the incorporation of more current timestep information as \textbf{memory} and the retention of more historical state information as \textbf{ignoring}. We define Mamba's Selectivity as the ability to prominently choose between memorizing and ignoring new timestep information under the current historical state, where it more prominently memorizes important timesteps while more prominently ignores noisy timesteps. As discussed in Section \ref{sec:mambablock}, Mamba's Selectivity arises from three key components: the matrices \textbf{A}, \textbf{B}, and the discretization parameter $\Delta$.

\subsection{Definition of a Selectivity Measurement}\label{measure}


Since Mamba, unlike Transformer, does not provide explicit attention scores to intuitively measure selectivity, understanding how it retains or discards information requires alternative strategies. While Mamba is based on a Selective State Space Model (SSM) rather than a gated recurrent mechanism, it shares with RNNs and LSTMs the key characteristic of processing sequences in a token-by-token manner, rather than consuming entire sequences simultaneously as Transformers do. This temporal nature of computation motivates us to draw inspiration from prior works analyzing memory and information retention in RNNs, where token-level dynamics — such as the evolution of hidden states across time — have been used to study memory behaviors \citep{zhang2020assessingmemoryabilityrecurrent,haviv2019understandingcontrollingmemoryrecurrent}.

Building on this perspective, we propose two quantitative metrics to assess the selectivity of Mamba. First, based on Eq. 4, we compute the correlation between the current hidden state \( h_t \), the previous hidden state \( h_{t-1} \), and the current input \( x_t \), normalizing the results to sum to 1. We define the correlation with \( x_t \) as the \textbf{memory score} \( s_t \) at the current time step. Following the definitions of memory and ignoring from Section 2.3, we categorize time steps with \( s_t > 0.7 \) as \textbf{Significant Memory (SM)}, those with \( s_t < 0.3 \) as \textbf{Significant Ignoring (SI)}, and the remainder as \textbf{Normal (NR)}. This approach is in line with previous efforts to quantify memory at each time step in sequential models by examining how hidden representations evolve over time \citep{8585721}. Based on these memory scores, we define two Selectivity Measurements: \textbf{Focus Ratio} and \textbf{Memory Entropy}.

1) Focus Ratio (FR): The proportion of Significant Memory and Significant Ignoring across all time steps. A higher FR indicates stronger selectivity.  
\begin{equation}
        FR = \frac{N_{SM} + N_{SI}}{N_{SM} + N_{SI} + N_{NR}}
\end{equation}

2) Memory Entropy (ME): The entropy of all memory scores, where higher ME indicates stronger selectivity.

\section{Method}
\begin{figure*}[tbp]
    \centering
    \includegraphics[width=\linewidth]{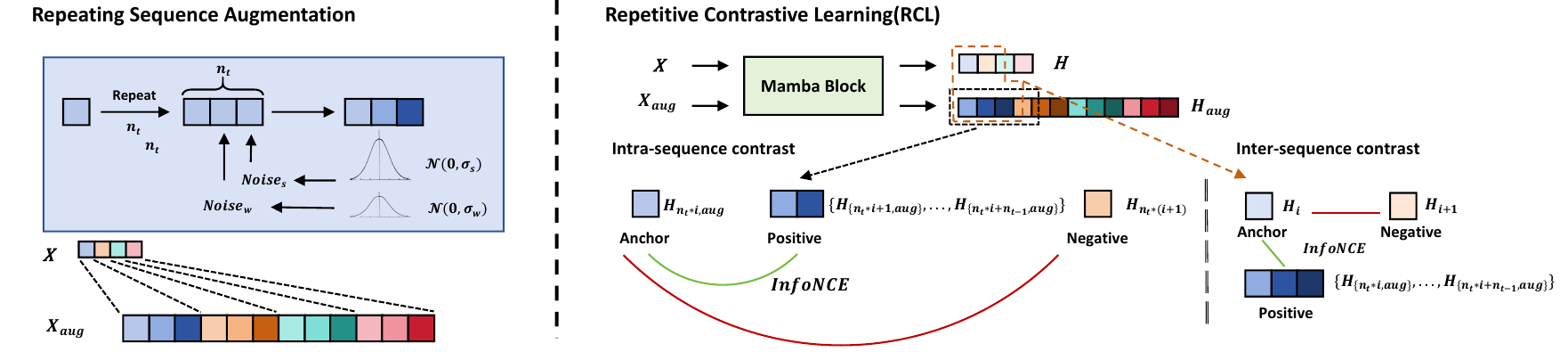}
    \caption{Process of the proposed method. Including Repeating Sequence Augmentation and Repetitive Contrastive Learning (RCL), with RCL consisting of Intra-sequence contrast and Inter-sequence contrast.}
    \label{fig:m1}
\end{figure*}

The Repetitive Contrastive Learning (RCL) paradigm is a pretraining method used before training the backbone model. It enhances the Mamba block's selective capabilities through initialization parameters, improving the backbone model's performance. RCL consists of three main steps. First, augmented data is created by repeating time steps and adding increasing noise, with positive and negative sample pairs defined at the time-step level. Second, intra-sequence and inter-sequence contrastive learning is applied to a single Mamba block. Intra-sequence learning helps the model ignore noisy time steps and focus on meaningful ones, while inter-sequence learning ensures robust temporal feature modeling and consistency across sequences of different lengths. Finally, the pretrained parameters are used to initialize all Mamba blocks in the backbone model. RCL only requires pretraining once on a single Mamba block but can be applied universally to all Mamba-based models as an initialization strategy. This method is efficient and scalable, adding no extra memory overhead and minimal time cost.

\subsection{Repeating Sequence Augmentation}
One significant reason why Mamba performs exceptionally well in time series prediction tasks is its selective structure. To enhance the selection capability of the Mamba Block, we designed the Repeating Sequence Augmentation. Specifically, as shown in Fig. \ref{fig:m1}, for each time step in each time series, we sequentially repeat this time step with repetition count $n_t$.

\begin{equation}
    \begin{aligned}
                        \textbf{X}_i \xrightarrow{\text{repeat}}& \ \textbf{X}_{i,1},..., \textbf{X}_{i,n_t}\\
    \textbf{X}_{\text{rep}} = \{\textbf{X}_{1,1},..., \textbf{X}_{1,n_t},& ..., \textbf{X}_{i,1},..., \\
    & \textbf{X}_{i,n_t}, ..., \textbf{X}_{s,1}, ..., \textbf{X}_{s,n_t} \}
    \end{aligned}
\end{equation}



where $\textbf{X}_i$ is the $i$-th step in time sequence, and $s$ is the length of the sequence. For the time series $\textbf{X} \in \mathbb{R}^{T \times F}$, $s = T$, the corresponding $\textbf{X}_{\text{rep}} \in \mathbb{R}^{(n_t*T) \times F}$. As for inverted time series $\textbf{X}^I \in \mathbb{R}^{F \times T}$, $s = F$, the corresponding $\textbf{X}^I_{\text{rep}} \in \mathbb{R}^{(n_t*F) \times T}$.

Then, we add Gaussian noise of increasing intensity, from weak to strong, to the repeated time steps. In our experiments, we choose $n_t = 3$, each time step $X_i$ is repeated and obtain $\textbf{X}_{i,1}, \textbf{X}_{i,2}, \textbf{X}_{i,2}$. We then sample a strong Gaussian noise and a weak Gaussian noise, and add them to the repeated time steps in increasing order of intensity, from weak to strong.

\begin{equation}
    \begin{aligned}
        \text{Noise}_\alpha &\sim \mathcal{N}(0, \sigma_\alpha^2)\\ 
        \text{Noise}_\beta &\sim \mathcal{N}(0, \sigma_\beta^2)\\
        \sigma_\alpha &< \sigma_\beta \\        
        \hat{\textbf{X}}_{i,2} &= \textbf{X}_{i,2} + \text{Noise}_\alpha\\
        \hat{\textbf{X}}_{i,3} &= \textbf{X}_{i,3} + \text{Noise}_\beta\\
        \textbf{X}_{\text{aug},i} &= \{\textbf{X}_{i,1}, \hat{\textbf{X}}_{i,2}, \hat{\textbf{X}}_{i,3}\} \\
        \textbf{X}_{\text{aug}} &= \textbf{X}_{\text{aug},1}\Vert \textbf{X}_{\text{aug},2}\Vert \ldots \Vert \textbf{X}_{\text{aug},s}
    \end{aligned}
\end{equation}

where $\text{Noise}_\alpha$ and $\text{Noise}_\beta$ represent weak and strong Gaussian noise, controlled by the variances $\sigma_\alpha$ and $\sigma_\beta$, $||$ denotes the sequential concatenation of sequences. Since the impact of noise accumulates progressively during sequence modeling, gradually increasing the noise effectively amplifies the distance between time steps. As a result, the repeated time steps form a sequence of denoising targets with progressively increasing difficulty.

\subsection{Repetitive Contrastive Learning}
We input both the original sequence $\textbf{X}$ and its augmented version $\textbf{X}_{\text{aug}}$ into the same Mamba Block, comparing their respective outputs $\textbf{H}$ and $\textbf{H}_{\text{aug}}$ to evaluate the Mamba Block's modeling capabilities across both sequences. As illustrated in Fig.\ref{fig:m1}, Repetitive Contrast Learning (RCL) encompasses two types of comparisons: intra-sequence contrast and inter-sequence contrast. Firstly, we define the output at any time step \( i \) with a repetition count \( n_t \) of the original sequence \( \textbf{X}_i \) as \( \textbf{H}_i \), and the output at the subsequent time step as \( \textbf{H}_{i+1} \). The outputs of the augmented sequence are represented as \( \{\textbf{H}_{\{i \cdot n_t, \text{aug}\}}, \textbf{H}_{\{i \cdot n_t + 1, \text{aug}\}}, \ldots, \textbf{H}_{\{i \cdot n_t + n_t - 1, \text{aug}\}}\} \), while the output at the next time step is \( \{\textbf{H}_{\{(i+1) \cdot n_t, \text{aug}\}}, \textbf{H}_{\{(i+1) \cdot n_t + 1, \text{aug}\}}, \ldots, \textbf{H}_{\{(i+1) \cdot n_t + n_t - 1, \text{aug}\}}\} \).

\textbf{Intra-sequence contrast} We hypothesize that if the Mamba Block possesses strong sequence selection capabilities, then the outputs $\{\textbf{H}_{\{i \cdot n_t, \text{aug}\}}, \textbf{H}_{\{i \cdot n_t + 1, \text{aug}\}}, \ldots, \textbf{H}_{\{i \cdot n_t + n_t - 1, \text{aug}\}}\}$ of the augmented sequence at the same time step should exhibit high similarity, while ignoring progressively increasing noise. Conversely, the outputs $\textbf{H}_{\{i \cdot n_t, \text{aug}\}}$ at the current time step and $\textbf{H}_{\{(i+1) \cdot n_t, \text{aug}\}}$ at the subsequent time step should have low similarity. Therefore, we define outputs at the same time step as positive examples, while outputs at the current and subsequent time steps serve as negative examples. The objective is to minimize the distance between positive examples and maximize the distance between negative examples within the sequence, thereby enhancing the Mamba Block's sequence selection capabilities. Specifically, we use $\textbf{H}_{\{i \cdot n_t, \text{aug}\}}$ as an anchor to form $n_t - 1$ positive samples and one negative sample, measuring similarity between samples using cosine similarity and employing the InfoNCE loss function \cite{oord2018representation}.

\begin{equation}
\begin{aligned}    \mathcal{L}_{\text{Intra}} =& - \frac{1}{s-1} \sum_{i=0}^{s-2} \log \frac{\sum_{z=1}^{n_t - 1}\exp(\text{sim}(\textbf{H}_{\{i \cdot n_t, \text{aug}\}}, \textbf{H}_{\{i \cdot n_t + z, \text{aug}\}}) / \tau)}{\sum_{z=1}^{n_t - 1}\exp(\text{sim}(\textbf{H}_{\{i \cdot n_t, \text{aug}\}}, \textbf{H}_{\{i \cdot n_t + z, \text{aug}\}}) / \tau)+\exp(\text{sim}(\textbf{H}_{\{i \cdot n_t, \text{aug}\}}, \textbf{H}_{\{(i+1) \cdot n_t, \text{aug}\}}) / \tau)}
\end{aligned}
\end{equation}

where $s$ is the sequence length, $i$ is the time step index, $n_t$ is the repetition count, $\tau$ is a temperature coefficient controlling the distinction of negative samples, and $\text{sim}(\cdot, \cdot)$ denotes the cosine similarity function, defined as:

\begin{equation}
\text{sim}(h_i, h_j) = \frac{h_i \cdot h_j}{\|h_i\| \|h_j\|}
\end{equation}
Intra-sequence contrast allows the Mamba Block to disregard noisy, repetitive time steps while prioritizing meaningful and effective ones, thereby strengthening its selection capabilities and noise resilience.

\textbf{Inter-sequence contrast} The inter-sequence contrast further enhances contrastive learning effects while preserving selection capability and temporal correlations on the original sequence, ensuring that the Mamba Block does not overfit to augmented data. Here, $\{\textbf{H}_{\{i \cdot n_t, \text{aug}\}}, \textbf{H}_{\{i \cdot n_t + 1, \text{aug}\}}, \ldots, \textbf{H}_{\{i \cdot n_t + n_t - 1, \text{aug}\}}\}$ and $\textbf{H}_i$ are defined as positive samples since they both represent the same time step and should maintain consistency across different time series lengths. Simultaneously, $\textbf{H}_i$ and $\textbf{H}_{i+1}$ are defined as negative samples to maintain selection capability on the original sequence.

\begin{equation}
\begin{aligned}
    \mathcal{L}_{\text{Inter}} = - \frac{1}{s-1} & \sum_{i=0}^{s-2} \log \frac{ \sum_{z=0}^{n_t - 1}\exp(\text{sim}(\textbf{H}_i, \textbf{H}_{\{i \cdot n_t + z, \text{aug}\}}) / \tau)}{\sum_{z=0}^{n_t - 1}\exp(\text{sim}(\textbf{H}_i, \textbf{H}_{\{i \cdot n_t + z, \text{aug}\}}) / \tau)+\exp(\text{sim}(\textbf{H}_i, \textbf{H}_{i+1}) / \tau)}
\end{aligned}
\end{equation}

where $s$, $i$, $n_t$, $\tau$, and $\text{sim}(\cdot, \cdot)$ are defined as above.

The overall optimization objective for Repetitive Contrastive Learning is:

\begin{equation}
    \mathcal{L}_{\text{rc}} = \mathcal{L}_{\text{Intra}} + \mathcal{L}_{\text{Inter}}
\end{equation}
It is noteworthy that the pre-training process for Repetitive Contrastive Learning is conducted exclusively on a single Mamba Block rather than the entire Mamba model. Even when sequence length is repeated, the memory usage and training time are typically lower than what is required for the entire model.


\subsection{Replace and Freezing Method}

After repetitive contrastive learning, we obtain Mamba block parameters with enhanced selective capabilities. We use these parameters as the initialization parameters for the Mamba blocks in various Mamba-based backbone models, replacing the original initialization method to improve the temporal prediction performance of them. 

Backbone models typically contain multiple Mamba blocks. We can choose to replace the initialization parameters for all blocks or only for a subset of them. The initialization parameters for other structures in the model, such as MLPs and attention mechanisms, remain unchanged. After parameter substitution, we can opt for full fine-tuning or partial fine-tuning of the replaced parameters. As discussed in Section \ref{mamba_select}, the selectivity of the Mamba block stems from the matrices $\textbf{A}$, $\textbf{B}$, and $\Delta$, where only $\textbf{A}$ is a globally optimizable matrix that encapsulates the common selective capabilities across all sequences. Therefore, in addition to full parameter fine-tuning, we recommend experimenting with freezing matrix $\textbf{A}$ to preserve the learned selective capabilities. 

Different parameter substitution and freezing methods may yield varying effects across different tasks. In Section \ref{subsec:Comparation}, we analyze the impact of different substitution ratios and freezing methods using a four-layer Mamba as an example.

\section{Experiment}

We conducted extensive experiments to validate the effectiveness of our method. In Section \ref{subsec:result}, we compare the prediction performance of various Mamba-based models—Mamba \citep{gu2024mambalineartimesequencemodeling}, iMamba, TimeMachine \citep{ahamed2024timemachinetimeseriesworth}, Bi-Mamba \citep{liang2024bimambabidirectionalmambatime} and SiMBA\citep{patro2024simbasimplifiedmambabasedarchitecture}—both with and without pre-trained parameters across multiple datasets: ETTh1, ETTh2, ETTm1, ETTm2, Traffic, and Electricity. 

In Section \ref{basicinfo}, we provide an overview of the experimental setup and basic information. In Section \ref{subsec:result}, we demonstrate the performance improvement of the backbone model achieved by RCL. In Section \ref{comp_pre}, we compare RCL with other pretraining methods. In Section \ref{subsec:ablation}, we present the ablation study results of RCL. In Section \ref{paramrcl}, we analyze the impact of RCL parameters. In Section \ref{comp_temporal}, we compare RCL with other temporal prediction models and achieve state-of-the-art results. In Section \ref{subsec:Comparation}, we examine the effects of different parameter substitution ratios and freezing methods. Additionally, in \ref{apdx_addre}, we list additional experimental results.

\subsection{Basic Information}\label{basicinfo}

\subsubsection{Mamba-based Baseline}
\begin{itemize} 
    \item \textbf{Mamba} \citep{gu2024mambalineartimesequencemodeling}: Mamba is a new Selective State Spaces model proposed by Albert Gu and Tri Dao in 2024.\citep{li2024mambandselectivestatespace} It demonstrates outstanding performance in sequence modeling through its selective state space formulation, effectively capturing long-range dependencies while maintaining computational efficiency. 
    \item \textbf{iMamba}: An enhancement of Mamba, iMamba builds upon the principles of the iTransformer, where features are treated as tokens. This model is tailored specifically for time series forecasting tasks, offering improved flexibility in feature tokenization. 
    \item \textbf{TimeMachine} \citep{ahamed2024timemachinetimeseriesworth}: TimeMachine, introduced by Md Atik Ahamed and Qiang Cheng in 2024, is designed for long-term sequence forecasting. By integrating channel-independent and channel-mixed modeling approaches, it achieves state-of-the-art performance. The architecture incorporates four Mamba blocks, optimizing predictive capability over extended sequences. 
    \item \textbf{Bi-Mamba} \citep{liang2024bimambabidirectionalmambatime}: Bi-Mamba was proposed in 2024, Bi-Mamba extends the Mamba framework by adaptively capturing both internal and inter-series dependencies in multivariate time series data. The model introduces forget gates, enabling it to retain relevant historical information over extended time periods, thereby enhancing its forecasting accuracy. 
    \item \textbf{SiMBA} 
    \citep{patro2024simbasimplifiedmambabasedarchitecture}: SiMBA is a hybrid architecture that combines Mamba-based sequence modeling with EinFFT, a novel FFT-based channel mixer. It is designed to overcome Mamba’s instability when scaling, offering a stable and efficient solution for large-scale sequence tasks. SiMBA refines selective state space models for improved scalability and performance in visual recognition. Due to space limitations, we omitted experiments whose output length longer than 96 for the SiMBA model.
    
\end{itemize}

\subsubsection{Temporal Baseline}
\begin{itemize}
    \item \textbf{Transformer}: \citep{vaswani2023attentionneed} The Transformer model, introduced by Vaswani et al. in 2017, revolutionized sequence modeling by using self-attention mechanisms. Its architecture allows for efficient parallelization and effectively captures long-range dependencies, making it highly suitable for various tasks such as natural language processing and time series forecasting.

    \item \textbf{iTransformer}:\citep{liu2024itransformerinvertedtransformerseffective} iTransformer is a restructured Transformer tailored for multivariate time series forecasting. Instead of embedding simultaneous time steps, it encodes each variate's full time series as a token, enabling better capture of global patterns and cross-variate correlations. This design aligns attention with the intrinsic structure of time series and achieves strong performance across forecasting benchmarks.
    
    \item \textbf{TimeMixer}: \citep{wang2024timemixerdecomposablemultiscalemixing} TimeMixer is a novel approach designed for time series modeling, leveraging the power of mixing operations to combine temporal features. By focusing on capturing intricate temporal dependencies and interactions, TimeMixer provides robust performance in both short-term and long-term forecasting tasks.
    
    \item \textbf{N-Beats}: \citep{oreshkin2020nbeatsneuralbasisexpansion} N-Beats is a deep learning architecture designed for univariate time series forecasting. It employs a \textit{Doubly Residual Stacking} mechanism, utilizing both backward and forward residual links to enhance signal propagation through a deep stack of fully connected layers. The model is highly flexible, requiring no time-series-specific components, and demonstrates state-of-the-art performance across diverse datasets, including M3, M4, and TOURISM. Additionally, N-Beats incorporates \textit{Ensembling} techniques during training, further improving its robustness and accuracy.

    \item \textbf{N-HiTS}: \citep{challu2022nhitsneuralhierarchicalinterpolation} N-HiTS builds upon N-Beats by introducing a \textit{Neural Basis Approximation Theorem} to enhance theoretical guarantees in forecasting. The model significantly improves long-horizon predictions through \textit{Multi-Rate Signal Sampling}, allowing it to focus on different frequency components dynamically. Additionally, N-HiTS employs a \textit{Hierarchical Interpolation} mechanism, which enables efficient decomposition and synthesis of forecasted signals, reducing volatility and computational complexity.

    \item \textbf{CrossFormer}: \citep{wang2021crossformerversatilevisiontransformer} CrossFormer introduces a cross-attention mechanism specifically tailored for time series data. It excels in integrating multiple time series inputs, enabling the model to learn complex relationships across different temporal sequences, thus improving forecasting accuracy and adaptability to diverse datasets.
    
    \item \textbf{PatchTST}: \citep{nie2023timeseriesworth64} PatchTST is a model that applies the concept of patch-based processing from computer vision to time series data. By segmenting time series into patches and processing them independently, PatchTST enhances the model's ability to capture local temporal patterns, improving efficiency and scalability for large datasets.
    
    \item \textbf{TimesNet}: \citep{wu2023timesnettemporal2dvariationmodeling} TimesNet is an advanced time series network that leverages a hierarchical structure to model temporal dependencies at multiple scales. This architecture allows TimesNet to adaptively focus on different temporal resolutions, providing superior performance in multiscale time series forecasting.
    
    \item \textbf{FEDFormer}: \citep{zhou2022fedformerfrequencyenhanceddecomposed} FEDFormer incorporates federated learning principles into the Transformer framework, allowing for decentralized time series modeling. This model is particularly effective in scenarios where data privacy is crucial, as it can learn from distributed data sources without centralizing the datasets.
    
    \item \textbf{Informer}: \citep{zhou2021informerefficienttransformerlong} Informer is designed to efficiently handle long sequences in time series forecasting. It introduces a ProbSparse self-attention mechanism that reduces computational complexity and memory usage, making it ideal for real-time applications and large-scale datasets. Informer achieves state-of-the-art results by focusing on significant temporal patterns while filtering out noise.
    
\end{itemize}

\subsubsection{Temporal Pre-training Baseline}

\begin{itemize}
    \item \textbf{SoftCLT}: \citep{softcls} SoftCLT is a cutting-edge model designed for contextual sequence learning. By incorporating soft clustering techniques, SoftCLT dynamically groups similar temporal patterns, enhancing the model's ability to generalize across varied contexts. This approach ensures superior performance in complex classification tasks, offering robust adaptability to fluctuating sequences while maintaining high interpretability.
    
    \item \textbf{InfoTS}: \citep{infots} InfoTS leverages information-theoretic principles to optimize time series modeling. By prioritizing the retention of informative features and minimizing redundancy, InfoTS significantly enhances predictive accuracy. This model excels in both supervised and unsupervised learning scenarios, making it versatile for diverse applications such as anomaly detection and trend analysis.
\end{itemize}

\subsubsection{Dataset}
Frequency, number of features, and time point information of the datasets.

\begin{table}[!ht]
    \centering
    \begin{tabular}{c|c|c|c|c}
    \hline
    \hline
        Dataset & Frequency & Features & Time Points & Split  \\ \hline
        ETTh1 & Hour & 7  & 17420  & 60\%/20\%/20\% \\ 
        ETTh2 & Hour & 7  & 17420  & 60\%/20\%/20\% \\ 
        ETTm1 & 15 Minutes & 7  & 69680  & 60\%/20\%/20\% \\ 
        ETTm2 & 15 Minutes & 7  & 69680  & 60\%/20\%/20\% \\ 
        Traffic & Hour & 862  & 17544 & 60\%/20\%/20\%  \\ 
        Electricity & Hour & 321  & 26304 & 60\%/20\%/20\%  \\ \hline\hline
    \end{tabular}
\end{table}

\subsubsection{Metric}
Mean Absolute Error (MAE):
$$
    \text{MAE} = \frac{1}{n} \sum_{i=1}^{n} \left| y_i - \hat{y}_i \right|
$$

Mean Squared Error (MSE):
$$
    \text{MSE} = \frac{1}{n} \sum_{i=1}^{n} \left( y_i - \hat{y}_i \right)^2
$$
\subsubsection{Model Settings}

The parameter settings for the Mamba block during pre-training are as follows: The model dimension ($d_{model}$) is set to values [16, 32, 64], and the state dimension ($d_{state}$) is set to [16, 64, 128]. The convolution dimension ($d_{conv}$) is fixed at 4, and $pad\_vocab\_size\_multiple$ is set to 8 to ensure consistent padding sizes. The expansion factor ($expand$) is configured to 2, with $conv\_bias$ enabled (set to True) and $bias$ disabled (set to False). The repeat time, denoted as $n_t$, is set to 3, while noise variance is varied between [0.001, 0.01]. During the inference phase, the Mamba Selective State Space Model (SSM) parameters are aligned with the corresponding pre-trained block parameters to maintain consistency and leverage learned patterns effectively. When comparing with other models and pre-training methods, we uniformly set $d_{model}=64$ and $d_{state}=16$. For the parameters of RCL, we choose to repeat each time step $n_t=3$ times, and the noise distribution is set to $\sigma_a=1e-3, \sigma_b=1e-2$.

\subsubsection{Training Settings}

The experiment was conducted on a server equipped with four NVIDIA GeForce RTX 3090 GPUs and an AMD EPYC 7282 16-Core Processor. We conducted each experiment at least three times and provided the standard deviation of the results in \ref{std}. During the pre-training phase, the number of layers ($n\_layer$) is set to 1, the number of epochs ($epoch$) is 100, the learning rate ($lr$) is configured to 1e-4, and the regularization coefficient is also set to 1e-4. In the inference stage, the maximum number of training epochs remains at 100, while $n\_layer$ is increased to 4. The Mean Absolute Error (MAE) serves as the loss function, and model selection is based on the lowest validation set loss. The parameter $frozentype$ is chosen as needed from the options [None, FrozenA], and the number of layers used for parameter replacement is selected from [25\%, 50\%, 75\%, 100\%], according to the specific experimental requirements. For the prediction length, we selected four different lengths: [96, 192, 336, 720] and conducted a series of experiments. However, all results tables presented in our paper, unless otherwise specified, use a prediction length of 96. This length was chosen because it effectively illustrates the corresponding conclusions and provides a clear basis for our findings.

\subsection{Main Result}
\label{subsec:result}

\begin{table*}[htbp]
  \centering

  \resizebox{1\linewidth}{!}{

        \begin{tabular}{c|c|cc|cc|cc|cc|cc|cc}
        \toprule
        \multicolumn{1}{c}{} &       & \multicolumn{2}{c|}{ETTh1} & \multicolumn{2}{c|}{ETTh2} & \multicolumn{2}{c|}{ETTm1} & \multicolumn{2}{c|}{ETTm2} & \multicolumn{2}{c|}{Traffic} & \multicolumn{2}{c}{Electricity} \\
    \cmidrule{3-14}    \multicolumn{1}{c}{} &       & MAE   & MSE   & MAE   & MSE   & MAE   & MSE   & MAE   & MSE   & MAE   & MSE   & MAE   & MSE \\
        \midrule
        \multirow{3}[2]{*}{Mamba} & w/o   & 0.6546  & 0.7672  & 1.4013  & 2.8442  & 0.5053  & 0.5432  & 0.5763  & 0.6008  & 0.4939  & 1.0279  & 0.4232  & 0.3926  \\
              & w     & 0.5974  & 0.6542  & 1.1536  & 2.0506  & 0.4798  & 0.4946  & 0.5646  & 0.5677  & 0.4604  & 0.9076  & 0.4168  & 0.3879  \\
              & up-rate\% & \textbf{8.7382 } & \textbf{14.729 } & \textbf{17.676 } & \textbf{27.902 } & \textbf{5.0465 } & \textbf{8.9470 } & \textbf{2.0302 } & \textbf{5.5093 } & \textbf{6.7827 } & \textbf{11.704 } & \textbf{1.5123 } & \textbf{1.1971 } \\
        \midrule
        \multirow{3}[2]{*}{iMamba} & w/o   & 0.4987  & 0.4928  & 0.6926  & 0.9084  & 0.4316  & 0.3998  & 0.4160  & 0.3666  & 0.3234  & 0.6538  & 0.2627  & 0.1857  \\
              & w     & 0.4472  & 0.4278  & 0.6833  & 0.8595  & 0.3970  & 0.3669  & 0.3304  & 0.2469  & 0.2913  & 0.6003  & 0.2597  & 0.1827  \\
              & up-rate\% & \textbf{10.327 } & \textbf{13.190 } & \textbf{1.3428 } & \textbf{5.3831 } & \textbf{8.0167 } & \textbf{8.2291 } & \textbf{20.577 } & \textbf{32.651 } & \textbf{9.9258 } & \textbf{8.1829 } & \textbf{1.1420 } & \textbf{1.6155 } \\
        \midrule
        \multirow{3}[2]{*}{TimeMachine} & w/o   & 0.3905  & 0.3833  & 0.3344  & 0.2911  & 0.3606  & 0.3342  & 0.2525  & 0.1746  & 0.3064  & 0.4983  & 0.2611  & 0.1872  \\
              & w     & 0.3869  & 0.3787  & 0.3298  & 0.2822  & 0.3458  & 0.3179  & 0.2508  & 0.1731  & 0.2991  & 0.4844  & 0.2586  & 0.1826  \\
              & up-rate\% & \textbf{0.9219 } & \textbf{1.2001 } & \textbf{1.3756 } & \textbf{3.0574 } & \textbf{4.1043 } & \textbf{4.8773 } & \textbf{0.6733 } & \textbf{0.8591 } & \textbf{2.3825 } & \textbf{2.7895 } & \textbf{0.9575 } & \textbf{2.4573 } \\
        \midrule
        \multirow{3}[2]{*}{Bi-Mamba} & w/o   & 0.3948  & 0.3813  & 0.3494 & 0.3073  & 0.3641& 0.3319    &   0.2704    &   0.1883    &   0.2786    &   0.587    &   0.2629    & 0.185 \\
              & w     & 0.3893  & 0.3794  & 0.3472 & 0.3  & 0.3578 & 0.3316   &   0.2707    &   0.1857    &   0.2761    &   0.5787    &   0.2611    & 0.1818 \\
              & up-rate\% & \textbf{1.3931 } & \textbf{0.4983 } & \textbf{0.6297} & \textbf{2.3755} & \textbf{1.7302} & \textbf{0.0903} & -0.1109 & \textbf{1.3808 } & \textbf{0.8973 } & \textbf{1.4140 } & \textbf{0.6847 } & \textbf{1.7280 } \\
        \midrule
        \multirow{3}[2]{*}{SiMBA} & w/o   & 0.4206 & 0.4033 & 0.4097 &0.3643 & 0.3841 &0.3466 & 0.2801 &0.1900 & 0.2601	& 0.5416 & 0.2433 &0.1531 \\
              & w     & 0.4109&0.3899&	0.3817&0.3238&	0.3742&0.3391&	0.2764&0.1868& 0.2566&0.5404 &	0.2412&0.1527   \\
              & up-rate\% & \textbf{2.2966}&\textbf{3.3153}&\textbf{6.8481}&\textbf{11.136}&\textbf{2.5720}&\textbf{2.1853}&\textbf{1.3149}&\textbf{1.6561}& \textbf{1.3315} & \textbf{0.2233} &\textbf{0.8496}&\textbf{0.3117} \\
              
        \bottomrule
        \end{tabular}
    
        }
    \caption{Comparison of performance improvement by replacing parameters obtained by RCL. w/o denotes no parameter replacement, w denotes parameter replacement, and up-rate represents the improvement rate.}
  \label{tab:mainresult}%
\end{table*}


We validated the performance improvements brought by the parameters of the pre-trained Mamba block across multiple Mamba-based models, as shown in Table \ref{tab:mainresult}. By leveraging the pre-trained Mamba block parameters, the Mamba model demonstrated substantial gains across various datasets, with the Mean Squared Error (MSE) reduced by up to 27.9\% and the Mean Absolute Error (MAE) improved by up to 17.7\%, averaging an improvement of over 5\%. For the iMamba model, the MAE showed gains of up to 20.6\%, while the MSE improved by up to 32.7\%, with an average performance increase exceeding 8\%. These results indicate that the Mamba block parameters, refined through Repetitive Contrastive Learning, significantly enhance the predictive capabilities of the Mamba and iMamba models in time series tasks, yielding average improvements of 5\% to 8\%.

For the TimeMachine model, MSE improved by up to 4.88\% and MAE by up to 4.10\%, with an average improvement of 2\%. While these gains are smaller compared to the Mamba and iMamba models, they remain noteworthy given that Bi-Mamba, SiMBA and TimeMachine are already state-of-the-art models for long-term sequence prediction. Achieving an additional 1\% to 2\% improvement solely by replacing the Mamba block parameters represents a meaningful advancement.

In summary, the parameters of the Mamba block, learned through the Repetitive Contrastive Learning method, consistently enhance the performance of various Mamba-based models. This underscores our method's efficacy in improving the sequence selection capability of the Mamba block and highlights its adaptability and potential for broad application.

\subsection{Comparison with Pre-training Methods}\label{comp_pre}

To systematically evaluate the latest pretraining methods in the time series domain, we conducted experiments on TS2Vec \citep{yue2022ts2vecuniversalrepresentationtime} and InfoTS \citep{infots}, and SoftCLT\citep{softcls}, with results summarized in Table \ref{tab:pretraincomp}. These experiments utilized official code implementations and focused on multivariate forecasting tasks. Below, we compare these contrastive methods based on two key dimensions: design goals, and experiments performance, contrasting these with our proposed pretraining approach tailored for Mamba-based models.

TS2Vec\citep{yue2022ts2vecuniversalrepresentationtime}, InfoTS\citep{infots}, and SoftCLT\cite{softcls} aim to enhance time series representation learning via contrastive pretraining, focusing on robust feature extraction for classification tasks. TS2Vec\citep{yue2022ts2vecuniversalrepresentationtime} employs hierarchical, dual-objective contrastive learning for generalizable embeddings, InfoTS\citep{infots} maximizes mutual information for discriminative representations, and SoftCLT\citep{softcls} leverages soft contrastive objectives to improve feature robustness. However, these methods, optimized for classification, rely on linear models for forecasting, limiting their efficacy in multivariate settings. In contrast, our RCL pretraining for Mamba-based models targets forecasting by learning intrinsic temporal patterns, enabling direct integration of pretrained parameters and achieving superior performance in forecasting tasks.

Our RCL pretraining approach, when applied to Mamba-based models, achieves a performance improvement of approximately 20\% in time series forecasting tasks compared to Mamba models pretrained with InfoTS, TS2Vec, or SoftCLT, as demonstrated by experimental results in Table \ref{tab:pretraincomp}.Unlike existing methods such as TS2Vec \citep{yue2022ts2vecuniversalrepresentationtime} and InfoTS \citep{infots}, which exhibit degraded performance when adapted to Mamba architectures due to their reliance on structurally incompatible feature extractors like TSEncoder, our method innovatively aligns pretraining with the linear processing strengths of Mamba models. By focusing on learning intrinsic temporal sampling rules and historical patterns through tailored contrastive learning, our approach ensures robust parameter initialization that directly enhances forecasting accuracy. This novel design not only outperforms alternative pretraining strategies but also establishes a new benchmark for scalable and effective time series forecasting, particularly in multivariate settings.

\begin{table*}[t]
  \centering

  \resizebox{\textwidth}{!}{
        \begin{tabular}{c|c|cc|cc|cc|cc|cc|cc|cc|cc|cc|cc}
    \toprule
    \multicolumn{2}{c|}{Model} & \multicolumn{2}{c|}{\textbf{TimeMachine*}} & \multicolumn{2}{c|}{\textbf{Bi-Mamba*}} & \multicolumn{2}{c|}{\textbf{Mamba*}} & \multicolumn{2}{c|}{\textbf{iMamba*}} & \multicolumn{2}{c|}{TS2Vec(TSEncoder)} & \multicolumn{2}{c|}{InfoTS(TSEncoder)} & \multicolumn{2}{c|}{SoftCLS(TSEncoder)} & \multicolumn{2}{c|}{TS2Vec(Mamba)} & \multicolumn{2}{c|}{InfoTS(Mamba)} & \multicolumn{2}{c}{SoftCLS(Mamba)} \\
    \midrule
    \multicolumn{2}{c|}{Metric} & \textbf{MAE} & \textbf{MSE} & \textbf{MAE} & \textbf{MSE} & \textbf{MAE} & \textbf{MSE} & \textbf{MAE} & \textbf{MSE} & \textbf{MAE} & \textbf{MSE} & \textbf{MAE} & \textbf{MSE} & \textbf{MAE} & \textbf{MSE} & \textbf{MAE} & \textbf{MSE} & \textbf{MAE} & \multicolumn{1}{c}{\textbf{MSE}} & \textbf{MAE} & \textbf{MSE} \\
    \midrule
    \multirow{4}[2]{*}{ETTh1} & 96    & \textbf{0.387 } & \textbf{0.379 } & \underline{0.389}  & \underline{0.379}  & 0.575  & 0.657  & 0.499  & 0.493  & 0.606  & 0.691  & 0.623  & 0.736  & 0.616  & 0.704  & 0.793  & 1.089  & 0.816  & 1.147  & 0.696  & 0.891  \\
          & 192   & \textbf{0.420 } & \underline{0.440}  & \underline{0.421}  & \textbf{0.425 } & 0.602  & 0.713  & 0.508  & 0.532  & 0.663  & 0.798  & 0.690  & 0.857  & 0.670  & 0.810  & 0.825  & 1.149  & 0.835  & 1.186  & 0.737  & 0.959  \\
          & 336   & \textbf{0.442 } & \underline{0.482}  & \underline{0.456}  & \textbf{0.481 } & 0.608  & 0.715  & 0.513  & 0.550  & 0.739 & 0.945 & 0.769  & 1.024  & 0.740  & 0.950  & 0.854  & 1.208  & 0.861  & 1.231  & 0.640  & 1.064  \\
\cmidrule{2-22}          & avg   & \textbf{0.416 } & \underline{0.434}  & \underline{0.422}  & \textbf{0.428 } & 0.595  & 0.695  & 0.506  & 0.525  & 0.669  & 0.811  & 0.694  & 0.872  & 0.676  & 0.821  & 0.824  & 1.149  & 0.838  & 1.188  & 0.691  & 0.971  \\
\midrule
    \multirow{4}[2]{*}{ETTh2} & 96    & \textbf{0.330 } & \textbf{0.282 } & \underline{0.347}  & \underline{0.300}  & 1.228  & 2.124  & 0.693  & 0.908  & 0.806  & 1.031  & 0.754  & 0.936  & 0.799  & 1.015  & 1.062  & 1.693  & 0.897  & 1.219  & 0.997  & 1.542  \\
          & 192   & \textbf{0.382 } & \textbf{0.355 } & \underline{0.394}  & \underline{0.373}  & 1.237  & 2.164  & 1.023  & 1.821  & 1.174  & 2.118  & 1.112  & 2.022  & 1.251  & 2.559  & 1.294  & 2.536  & 1.251  & 2.506  & 1.343  & 2.820  \\
          & 336   & \textbf{0.420 } & \textbf{0.412 } & \underline{0.429}  & \underline{0.434}  & 1.234  & 2.153  & 1.073  & 2.042  & 1.308  & 2.618  & 1.264  & 2.482  & 1.312  & 2.639  & 1.375  & 2.763  & 1.327  & 2.733  & 1.402  & 2.952  \\
\cmidrule{2-22}          & avg   & \textbf{0.377 } & \textbf{0.350 } & \underline{0.390}  & \underline{0.369}  & 1.233  & 2.147  & 0.929  & 1.590  & 1.096  & 1.922  & 1.043  & 1.813  & 1.121  & 2.071  & 1.244  & 2.331  & 1.158  & 2.152  & 1.248  & 2.438  \\
\midrule
    \multirow{4}[2]{*}{ETTm1} & 96    & \textbf{0.346 } & \textbf{0.318 } & \underline{0.358} & \underline{0.332}  & 0.492  & 0.528  & 0.432  & 0.400  & 0.530  & 0.572  & 0.540  & 0.602  & 0.534  & 0.581  & 0.774  & 1.031  & 0.741  & 0.985  & 0.623  & 0.808  \\
          & 192   & \textbf{0.377 } & \underline{0.375}  & \underline{0.384}  & \textbf{0.369 } & 0.513  & 0.587  & 0.450  & 0.439  & 0.562  & 0.624  & 0.575  & 0.649  & 0.569  & 0.635  & 0.784  & 1.051  & 0.756  & 1.014  & 0.654  & 0.849  \\
          & 336   & \textbf{0.387 } & \textbf{0.396 } & \underline{0.407}  & \underline{0.404}  & 0.817  & 1.457  & 0.491  & 0.509  & 0.603  & 0.686  & 0.622  & 0.729  & 0.610  & 0.697  & 0.802  & 1.088  & 0.770  & 1.040  & 0.681  & 0.885  \\
\cmidrule{2-22}          & avg   & \textbf{0.370 } & \textbf{0.363 } & \underline{0.383}  & \underline{0.368}  & 0.607  & 0.857  & 0.458  & 0.449  & 0.565  & 0.627  & 0.579  & 0.660  & 0.571  & 0.638  & 0.787  & 1.056  & 0.755  & 1.013  & 0.653  & 0.847  \\
\midrule
    \multirow{4}[2]{*}{ETTm2} & 96    & \textbf{0.251 } & \textbf{0.173 } & \underline{0.271}  & \underline{0.186}  & 0.576  & 0.601  & 0.416  & 0.367  & 0.453  & 0.401  & 0.452  & 0.377  & 0.460  & 0.400  & 0.855  & 1.113  & 0.782  & 0.969  & 0.491  & 0.437  \\
          & 192   & \textbf{0.293 } & \textbf{0.238 } & \underline{0.313}  & \underline{0.254}  & 0.667  & 0.847  & 0.497  & 0.495  & 0.576  & 0.579  & 0.560  & 0.542  & 0.580  & 0.587  & 0.923  & 1.282  & 0.857  & 1.152  & 0.591  & 0.590  \\
          & 336   & \textbf{0.333 } & \textbf{0.299 } & \underline{0.364}  & \underline{0.316}  & 0.705  & 0.922  & 0.793  & 1.032  & 0.717  & 0.862  & 0.713  & 0.846  & 0.730  & 0.885  & 1.008  & 1.540  & 0.969  & 1.461  & 0.730  & 0.855  \\
\cmidrule{2-22}          & avg   & \textbf{0.292 } & \textbf{0.237 } & \underline{0.316}  & \underline{0.252}  & 0.650  & 0.790  & 0.569  & 0.631  & 0.582  & 0.614  & 0.575  & 0.588  & 0.590  & 0.624  & 0.929  & 1.311  & 0.869  & 1.194  & 0.604  & 0.627  \\
\midrule
    \multirow{4}[3]{*}{Electricity} & 96    & \textbf{0.259 } & \underline{0.183}  & \underline{0.261}  & \textbf{0.182 } & 0.423  & 0.393  & 0.260  & 0.183  & 0.403  & 0.329  & 0.290  & 0.380  & 0.401  & 0.326  & 0.635  & 0.696  & 0.531  & 0.524  & 0.553  & 0.571  \\
          & 192   & \textbf{0.246 } & \textbf{0.152 } & \underline{0.270}  & \underline{0.188}  & 0.430  & 0.405  & 0.280  & 0.205  & 0.406  & 0.330  & 0.293  & 0.383  & 0.403  & 0.327  & 0.636  & 0.698  & 0.532  & 0.524  & 0.555  & 0.573  \\
          & 336   & \textbf{0.261 } & \textbf{0.169 } & \underline{0.283}  & \underline{0.200}  & 0.435  & 0.411  & 0.298  & 0.222  & 0.419  & 0.348  & 0.311  & 0.396  & 0.416  & 0.344  & 0.639  & 0.702  & 0.540  & 0.538  & 0.565  & 0.581  \\
\cmidrule{2-22}          & avg   & \textbf{0.255 } & \textbf{0.168 } & \underline{0.271}  & \underline{0.190}  & 0.429  & 0.403  & 0.279  & 0.203  & 0.409  & 0.335  & 0.298  & 0.386  & 0.407  & 0.332  & 0.637  & 0.699  & 0.534  & 0.529  & 0.558  & 0.575  \\
    \bottomrule
    \end{tabular}%
  }
  \caption{Comparison results with pre-training methods. Bolded names with an asterisk indicate models using our pre-training methods. Parentheses following InfoTS and SoftCLT denote the backbone models utilized during pre-training. The best results for each metric are highlighted in bold.}
  \label{tab:pretraincomp}%
\end{table*}%

\begin{table*}[htbp]
  \centering

  \resizebox{\textwidth}{!}{
    \begin{tabular}{c|cc|cc|cc|cc|cc|cc}
    \toprule
    \textbf{Model} 
    & \multicolumn{2}{c|}{\textbf{ETTh1}} 
    & \multicolumn{2}{c|}{\textbf{ETTh2}} 
    & \multicolumn{2}{c|}{\textbf{ETTm1}} 
    & \multicolumn{2}{c|}{\textbf{ETTm2}} 
    & \multicolumn{2}{c|}{\textbf{Traffic}} 
    & \multicolumn{2}{c}{\textbf{Electricity}} \\
    \midrule
    Metric & MAE & MSE & MAE & MSE & MAE & MSE & MAE & MSE & MAE & MSE & MAE & MSE \\
    \midrule
    TimeMachine* & \textbf{0.429} & 0.447 & \textbf{0.390} & \textbf{0.365} & \textbf{0.385} & \underline{0.386} & \textbf{0.317} & \underline{0.278} & \underline{0.287} & \underline{0.446} & \textbf{0.265} & \textbf{0.176} \\
    TimeMachine   & \underline{0.432} & 0.452 & 0.397 & \underline{0.376} & \underline{0.391} & 0.395 & \underline{0.319} & 0.282 & 0.290 & 0.454 & \underline{0.269} & 0.181 \\
    Bi-Mamba*     & 0.441 & \underline{0.445} & 0.443 & 0.460 & 0.398 & 0.391 & 0.340 & 0.290 & 0.308 & 0.640 & 0.283 & 0.206 \\
    Bi-Mamba      & 0.445 & 0.452 & 0.445 & 0.459 & 0.404 & 0.395 & 0.351 & 0.317 & 0.307 & 0.644 & 0.287 & 0.212 \\
    iTransformer  & 0.448 & 0.454 & 0.407 & 0.383 & 0.410 & 0.407 & 0.332 & 0.288 & \textbf{0.282} & \textbf{0.428} & 0.270 & \underline{0.178} \\
    TimeMixer     & 0.440 & 0.447 & \underline{0.395} & 0.365 & 0.396 & \textbf{0.381} & 0.323 & \textbf{0.275} & 0.298 & 0.485 & 0.273 & 0.182 \\
    CrossFormer   & 0.522 & 0.529 & 0.684 & 0.942 & 0.495 & 0.513 & 0.611 & 0.757 & 0.304 & 0.550 & 0.334 & 0.244 \\
    PatchTST      & 0.455 & 0.469 & 0.407 & 0.387 & 0.400 & 0.387 & 0.326 & 0.281 & 0.362 & 0.555 & 0.304 & 0.216 \\
    TimesNet      & 0.450 & 0.458 & 0.427 & 0.414 & 0.406 & 0.400 & 0.333 & 0.291 & 0.336 & 0.620 & 0.295 & 0.193 \\
    FEDFormer     & 0.460 & \textbf{0.440} & 0.449 & 0.437 & 0.452 & 0.448 & 0.349 & 0.305 & 0.376 & 0.610 & 0.327 & 0.214 \\
    Informer      & 0.795 & 1.040 & 1.729 & 4.431 & 0.734 & 0.961 & 0.810 & 1.410 & 0.397 & 0.311 & 0.416 & 0.764 \\
    N-HiTS        & 0.455 & 0.475 & 0.448 & 0.421 & 0.416 & 0.410 & 0.330 & 0.279 & 0.311 & 0.452 & 0.329 & 0.246 \\
    N-BEATS       & 0.488 & 0.490 & 0.471 & 0.411 & 0.401 & 0.418 & 0.345 & 0.294 & 0.321 & 0.461 & 0.329 & 0.246 \\
    \bottomrule
    \end{tabular}

}
    \caption{Comparison results with temporal model. Bolded numbers indicate optimal results and underscores indicate sub-optimal results.}
    \label{tab:tcompmin}
\end{table*}

\subsection{Ablation Study}
\label{subsec:ablation}

\begin{table}[t]
  \centering
  \begin{minipage}[t]{0.48\textwidth}
        \centering
        \resizebox{\linewidth}{!}{
        \begin{tabular}{c|cc|cc}
        \toprule
        \multirow{2}[4]{*}{} & \multicolumn{2}{c|}{ETTh1} & \multicolumn{2}{c}{ETTh2} \\
    \cmidrule{2-5}          & MAE   & MSE   & MAE   & MSE \\
        \midrule
        w/o intra-sequence contrast & 0.636  & 0.743  & 1.351  & 2.659  \\
        w/o inter-sequence contrast & 0.622  & 0.710  & 1.296  & 2.421  \\
        w/o noise  & 0.655  & 0.767  & 1.401  & 2.844  \\
        our approach & \textbf{0.597 } & \textbf{0.654 } & \textbf{1.154 } & \textbf{2.051 } \\
        \bottomrule
        \end{tabular}%
      }
      \caption{Ablation results for our contrastive method settings, highlighting the effects of intra-sequence, inter-sequence, and noise augmentation components, which correspond to the three key parts of our model design.}
      \label{tab:ab1}%
  \end{minipage}%
  \hfill
  \begin{minipage}[t]{0.48\textwidth}
      \centering
      \resizebox{\linewidth}{!}{
        \begin{tabular}{c|cc|cc}
        \toprule
        \multirow{2}[4]{*}{} & \multicolumn{2}{c|}{ETTh1} & \multicolumn{2}{c}{ETTh2} \\
    \cmidrule{2-5}          & MAE   & MSE   & MAE   & MSE \\
        \midrule
        RCL w uniform noise & 0.601  & 0.664  & 1.158  & 2.060  \\
        RCL w constant-intensity Gaussian noise & 0.600  & 0.660  & 1.155  & 2.059  \\
        RCL w increasing-intensity Gaussian noise & \textbf{0.597 } & \textbf{0.654 } & \textbf{1.154 } & \textbf{2.051 } \\
        \bottomrule
        \end{tabular}%
      }
      \caption{Ablation study on the design of increasing-intensity Gaussian noise. We conducted a series of explorations examining different noise formats and their impact.}
      \label{tab:ab2}%
  \end{minipage}
\end{table}

We conducted two ablation experiments to evaluate our proposed RCL method. All ablation experiments used a 4-layer Mamba as the baseline model. In the first ablation experiment, as shown in Table \ref{tab:ab1}, we separately removed intra-sequence contrast, inter-sequence contrast, and noise. Removing intra-sequence contrast significantly reduced prediction performance because this contrast enhances the Mamba block's ability to select time steps and denoise. Without it, the model's ability to select time steps diminishes. Similarly, removing inter-sequence contrast also led to performance loss, as repeated time sequences can disrupt temporal consistency. The purpose of inter-sequence contrast is to maintain consistency with the temporal features of the original sequence. Without it, RCL cannot learn temporal features in broken sequences. The most significant performance drop occurred when noise was removed. Without added noise, repeated time steps are indistinguishable from the original ones, reducing task difficulty and failing to enhance the Mamba block's ability to resist noise and select time steps.

In the second ablation experiment, as shown in Table \ref{tab:ab2}, we compared the effects of different types of noise on performance. Specifically, we compared uniform noise, constant-intensity Gaussian noise, and increasing-intensity Gaussian noise used in RCL. All three types of noise yielded good results, with uniform noise performing slightly worse than constant-intensity Gaussian noise, and constant-intensity Gaussian noise performing slightly worse than increasing-intensity Gaussian noise. The increasing-intensity Gaussian noise further accentuates differences between repeated time steps, increasing the difficulty of distinguishing effective information from noise, thereby enhancing pre-training performance.
\subsection{Hyper-Parameter Experiment of RCL}\label{paramrcl}
\begin{table}[ht]

    \centering
    \begin{minipage}[t]{0.5\linewidth}
    \centering
    \begin{minipage}[t]{0.5\linewidth}
        \resizebox{\linewidth}{!}{
        \begin{tabular}{c|cc|cc}
            \toprule
            \multirow{2}[4]{*}{$n_t$} & \multicolumn{2}{c|}{ETTh1} & \multicolumn{2}{c}{ETTh2} \\
        \cmidrule{2-5}          & MAE   & MSE   & MAE   & MSE \\
            \midrule
            2     & 0.623  & 0.708  & 1.182  & 2.145  \\
            3     & \textbf{0.597}  & \textbf{0.654}  & \textbf{1.154}  & \textbf{2.051}  \\
            4     & 0.591  & 0.653  & 1.148  & 2.003  \\
            \bottomrule
        
        \end{tabular}%
    }
    \end{minipage}
    \hfill
    \begin{minipage}[t]{0.45\linewidth}    
    \resizebox{\linewidth}{!}{  
        \begin{tabular}{c|cc|cc}
          \toprule
        \multirow{2}[4]{*}{$\sigma_a$} & \multicolumn{2}{c|}{ETTh1} & \multicolumn{2}{c}{ETTh2} \\
        \cmidrule{2-5}          & MAE   & MSE   & MAE   & MSE \\
            \midrule
            5e-4  & 0.614  & 0.671  & 1.180  & 2.085  \\
            1e-3  & \textbf{0.597}  & \textbf{0.654}  & \textbf{1.154}  & \textbf{2.051}  \\
            5e-3  & 0.601  & 0.700  & 1.172  & 2.072  \\
            1e-2  & 0.629  & 0.683  & 1.199  & 2.096  \\
        \bottomrule
        \end{tabular}%
    }
    \end{minipage}
    \label{tab:paramrcl}%
    \caption{RCL Parameter Experiment Results}
    \end{minipage}
    \begin{minipage}[t]{0.45\linewidth}
    \centering
        \centering
        \resizebox{\linewidth}{!}{  
        \begin{tabular}{c|ccc|cc}
        \hline
            ~ & SM & SI & NR & FR & ME  \\ 
            \hline
            w RCL & 11897 & 32789 & 219889 & 0.1689 & 1.53  \\ 
            w/o RCL & 5641 & 12875 & 246059 & 0.0700 & 1.04  \\ 
            \hline
        \end{tabular}
        \label{tab:frme}
    }
    
    \caption{ Focus Ratio and Memory Entropy}            
    \end{minipage}
\end{table}%


We conducted experimental comparisons on the RCL training parameters. Specifically, we compared different repetition counts \( n_t \) and initial noise intensities \( \sigma_a \). We set the noise intensity for each repetition is twice that of the previous one. The experimental results are shown in Table 5. It can be observed that when \( n_t = 2 \), the performance is significantly lower than others. This is because fewer repetitions make the task simpler and do not significantly enhance the original model. When \( n_t = 4 \), the performance is only slightly better than when \( n_t = 3 \). However, considering training time and memory usage, we believe that overall, \( n_t = 3 \) is preferable. Regarding the initial noise intensity \( \sigma_a \), when \( \sigma_a = 5 \times 10^{-3} \), the noise is relatively weak, causing low interference and making the task simpler, resulting in weaker performance improvement. When \( \sigma_a \) is greater than \( 1 \times 10^{-3} \), the noise becomes too strong, significantly differing from the original temporal signals, thus reducing the difficulty of recognition and leading to less performance improvement.

\subsection{Comparison with Temporal Model}\label{comp_temporal}
We compared our approach with existing state-of-the-art time series prediction models. We set all input lengths to 96 and conducted experiments across multiple prediction horizons $ T = \{96, 192, 336, 720\} $. The average results across the four prediction horizons are presented in Table \ref{tab:tcompmin}, while the detailed results for each individual prediction horizon are provided in Table \ref{tab:tcomp} of \ref{full_compare}. TimeMachine* and Bi-Mamba* refer to the TimeMachine and Bi-Mamba models initialized with parameters obtained using RCL. Our method achieves optimal results across various datasets and prediction horizons. For datasets with fewer data channels, our approach consistently achieves the best Mean Absolute Error (MAE) results across all prediction horizons, and Mean Squared Error (MSE) results are generally among the top two. For datasets with more channels, such as traffic and electricity, our method shows more significant improvements for longer prediction targets. This indicates enhanced stability in long-sequence predictions, attributed to the parameters obtained through RCL, which enable the Mamba block to have stronger selectivity for time series data.

\subsection{Comparison of Replacement and Freezing Methods}
\label{subsec:Comparation}

\begin{table*}[htbp]
  \centering
  \resizebox{1\linewidth}{!}{
        \begin{tabular}{c|c|cc|cc|cc|cc}
        \toprule
        \multicolumn{1}{c}{} &       & \multicolumn{4}{c|}{ETTm1}    & \multicolumn{4}{c}{ETTm2} \\
    \cmidrule{3-10}    \multicolumn{1}{c}{} &       & \multicolumn{2}{c|}{None} & \multicolumn{2}{c|}{FrozenA} & \multicolumn{2}{c|}{None} & \multicolumn{2}{c}{FrozenA} \\
    \cmidrule{3-10}    \multicolumn{1}{c}{} &       & MAE   & MSE   & MAE   & MSE   & MAE   & MSE   & MAE   & MSE \\
        \midrule
              & w/o   & 0.5053  & 0.5432  & 0.5053  & 0.5432  & 0.5763  & 0.6008  & 0.5763  & 0.6008  \\
        \midrule
        \multirow{2}[2]{*}{layer-25\%} & w     & 0.4921  & 0.5394  & 0.4921  & 0.5393  & 0.6609  & 0.7902  & 0.5611  & 0.5696  \\
              & up-rate\% & \textbf{2.6123 } & \textbf{0.6996 } & \textbf{2.6123 } & \textbf{0.7180 } & {-14.6799 } & {-31.5246 } & \textbf{2.6375 } & \textbf{5.1931 } \\
        \midrule
        \multirow{2}[1]{*}{layer-50\%} & w     & 0.4798  & 0.4946  & 0.4976  & 0.5548  & 0.6021  & 0.6230  & 0.6389  & 0.7423  \\
              & up-rate\% & \textbf{5.0465 } & \textbf{8.9470 } & \textbf{1.5238 } & {-2.1355 } & {-4.4768 } & {-3.6951 } & {-10.8624 } & {-23.5519 } \\
        \midrule
        \multirow{2}[1]{*}{layer-75\%} & w     & 0.4816  & 0.5256  & 0.4816  & 0.5255  & 0.5299  & 0.5366  & 0.5646  & 0.5676  \\
              & up-rate\% & \textbf{4.6903 } & \textbf{3.2401 } & \textbf{4.6903 } & \textbf{3.2585 } & \textbf{8.0514 } & \textbf{10.6858 } & \textbf{2.0302 } & \textbf{5.5260 } \\
        \midrule
        \multirow{2}[2]{*}{layer-100\%} & w     & 0.5106  & 0.5692  & 0.5016  & 0.5658  & 0.5486  & 0.5735  & 0.5296  & 0.5258  \\
              & up-rate\% & {-1.0489 } & {-4.7865 } & \textbf{0.7322 } & {-4.1605 } & \textbf{4.8065 } & \textbf{4.5439 } & \textbf{8.1034 } & \textbf{12.4834 } \\
        \bottomrule
        \end{tabular}%
    }
    \caption{Comparison of Replacement and Freezing Methods. The "layer-x\%" indicates that the first x\% of layers were replaced by pre-trained blocks.}
  \label{tab:randf}%
\end{table*}%

A Mamba-based model typically comprises multiple Mamba blocks. Each Mamba block contains a matrix $\textbf{A}$, which is defined in 3.2. The parameters are responsible for controlling the block’s selectivity towards information before. To evaluate the impact of parameter replacement and parameter freezing during the inference stage, we used a 4-layer Mamba model as a baseline. The replacement strategy involved substituting 25\%, 50\%, 75\%, and 100\% of the Mamba blocks, while the parameter freezing strategy was categorized into no freezing (None) and freezing of matrix $\textbf{A}$ (FrozenA). Freezing matrix $\textbf{A}$ helps preserve the enhanced selectivity gained during pre-training.

As shown in Table \ref{tab:randf}, the optimal parameter replacement and freezing strategies differ across datasets. For the ETTm1 dataset, replacing 50\% of the Mamba blocks without freezing any parameters yielded the greatest improvement, while replacing 100\% of the blocks resulted in the lowest performance. This suggests that the selection capabilities of the pre-trained parameters do not fully align with the prediction target. By replacing only 50\% of the Mamba blocks, the model can better encode the time series, while the remaining blocks focus on fitting the specific prediction requirements of the dataset, ultimately enhancing model performance.

Conversely, for the ETTm2 dataset, the greatest improvement was achieved by replacing all Mamba blocks and freezing matrix $\textbf{A}$. In this case, the selective enhancements from pre-training aligned well with the dataset's prediction targets. This approach preserved the pre-trained parameters' selectivity while allowing the remaining parameters to adjust to fit the prediction targets effectively.

Similar results were observed across other datasets. Broadly, the findings can be grouped into two effective strategies: replacing 50\% of the Mamba blocks without freezing any parameters and replacing 100\% of the Mamba blocks while freezing matrix $\textbf{A}$. We recommend choosing between these two approaches during the inference phase for optimal performance.
\begin{figure}[ht]
    \centering
    \includegraphics[width=\linewidth]{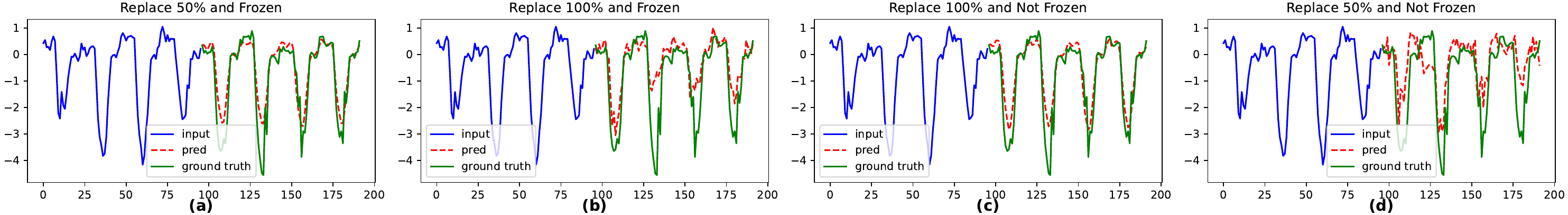}
    \caption{Visualization of Predictions when Replacing 100\%/50\% of the Layers and Frozen/Not Forzen Parameters}
    \label{fig:virturl_replace}
\end{figure}

We visualized the predictions under different settings of replacing 100\%/50\% of the parameters and whether to freeze them, as shown in Fig. \ref{fig:virturl_replace}, providing a more intuitive understanding of the impact of these configurations. Overall, the performance of replacing 100\% + Not Frozen is similar to replacing 50\% + Frozen, while replacing 100\% + Frozen and replacing 50\% + Not Frozen show poorer performance. This is because, for multi-layer models (whether Mamba or Transformer), deeper layers tend to model specificity, while shallower layers tend to model commonality. For time series, this can be intuitively understood as short-term fluctuations and long-term periodicity. From the practical results, comparing (a) and (b), replacing only 50\% retains sufficient parameters to focus on learning specific predictive features, whereas replacing 100\% leads to poorer specificity fitting and reduced ability to fit extreme values. Comparing (b) and (c), not freezing any parameters alleviates the issue of insufficient trainable parameters when replacing 100\%, achieving comparable performance. Comparing (a) and (d), when replacing fewer layers and not freezing parameters, the information learned by RCL can be lost, thus requiring freezing the matrix to retain the unique information learned during pre-training.

\subsection{Few-Shot Forecasting}
\noindent\textbf{Setup} To verify the performance of our method in few-shot scenarios, we designed the Few-Shot Forecasting experiment. Specifically, we retained only the first 10\% of the original dataset as the training set to pre-train the Mamba block and train the backbone model, while keeping the size of the test set unchanged. In this setup, not only is the amount of training data significantly reduced, but there is also a substantial temporal gap between the test set and the training set, further increasing the difficulty of the prediction task.
\begin{table}[htbp]
  \centering
    \resizebox{\linewidth}{!}{
    \begin{tabular}{c|cc|cc|cc|cc|cc|cc|cc|cc|cc}
    \toprule
    Model & \multicolumn{2}{c|}{\textbf{TimeMachine*}} & \multicolumn{2}{c|}{TimeMachine} & \multicolumn{2}{c|}{\textbf{Bi-Mamba*}} & \multicolumn{2}{c|}{Bi-Mamba} & \multicolumn{2}{c|}{iTransformer} & \multicolumn{2}{c|}{TimeMixer} & \multicolumn{2}{c|}{CrossFormer} & \multicolumn{2}{c|}{TimesNet} & \multicolumn{2}{c}{Informer} \\
    \midrule
    Metric & \textbf{MAE} & \textbf{MSE} & \textbf{MAE} & \textbf{MSE} & \textbf{MAE} & \textbf{MSE} & \textbf{MAE} & \textbf{MSE} & \textbf{MAE} & \textbf{MSE} & \textbf{MAE} & \textbf{MSE} & \textbf{MAE} & \textbf{MSE} & \textbf{MAE} & \textbf{MSE} & \textbf{MAE} & \textbf{MSE} \\
    \midrule
    \textbf{ETTh1} & \textbf{0.452 } & \textbf{0.471 } & 0.479  & 0.504  & 0.504  & \underline{0.476}  & \underline{0.464}  & 0.489  & 0.470  & 0.499  & 0.728  & 1.025  & 0.800  & 1.405  & 0.863  & 1.345  & 0.814  & 1.164  \\
    \textbf{ETTh2} & \underline{0.353}  & \textbf{0.319 } & 0.360  & 0.325  & \textbf{0.325 } & \underline{0.320}  & 0.369  & 0.335  & 0.363  & 0.323  & 1.165  & 2.692  & 1.216  & 2.588  & 1.976  & 5.966  & 1.149  & 2.264  \\
    \textbf{ETTm1} & \textbf{0.403 } & \textbf{0.419 } & 0.423  & 0.438  & 0.438  & \underline{0.426}  & \underline{0.417}  & 0.438  & 0.428  & 0.442  & 0.551  & 0.616  & 0.631  & 0.846  & 1.057  & 1.960  & 0.821  & 1.125  \\
    \textbf{ETTm2} & \underline{0.241}  & \textbf{0.171 } & 0.263  & 0.188  & \textbf{0.239 } & \underline{0.185}  & 0.266  & 0.191  & 0.271  & 0.194  & 0.400  & 0.315  & 0.896  & 1.497  & 1.347  & 3.058  & 1.284  & 2.763  \\
    \textbf{Traffic} & \underline{0.331}  & 0.517  & 0.341  & 0.534  & \textbf{0.328 } & \textbf{0.493 } & 0.335  & 0.505  & 0.332  & \underline{0.500}  & 0.566  & 1.040  & 0.463  & 0.892  & 0.443  & 0.888  & 0.754  & 1.453  \\
    \textbf{Electricity} & \underline{0.259}  & 0.193  & 0.277  & 0.199  & \textbf{0.251 } & \textbf{0.181 } & 0.268  & 0.189  & 0.264  & \underline{0.188}  & 0.409  & 0.333  & 0.363  & 0.278  & 0.600  & 0.707  & 0.871  & 1.255  \\
    \bottomrule
    \end{tabular}%
    }
  \caption{Comparative Results Under Few-Shot Setting. The best results are \textbf{bolded}, and the second-best results are \underline{underlined}.}
  \label{tab:fewshot}%
\end{table}%

\noindent\textbf{Result} The comparative results are shown in Table \ref{tab:fewshot}. \textbf{TimeMachine*} and \textbf{Bi-Mamba*} represent the RCL-enhanced versions of TimeMachine and Bi-Mamba, respectively. First, in the few-shot forecasting problem, the performance of all backbone models experienced a significant decline, indicating that the backbone models are heavily dependent on the amount of data. However, by comparing the results of TimeMachine and Bi-Mamba with and without RCL, it is evident that our method significantly enhances the performance of the backbone models, bringing their capabilities closer to the results achieved with the full dataset. This demonstrates that RCL can uncover more potential temporal patterns even with limited data, thereby reducing the backbone models' reliance on the volume of data.

\subsection{Long-Term Forecasting Under Missing Points Data}
\noindent\textbf{Setup} To validate the performance of our method under the condition of missing points, we conducted experiments with missing data. Specifically, we randomly masked 10\% of the input sequence and predicted future variations under the scenario of missing data.  Both the RCL and backbone model training utilized the same masking results. This experimental setup allows us to verify the performance of our method in the presence of missing data.

\begin{table}[htbp]
  \centering
    \resizebox{\linewidth}{!}{
    \begin{tabular}{c|cc|cc|cc|cc|cc|cc|cc|cc|cc}
    \toprule
    Model & \multicolumn{2}{c|}{\textbf{TimeMachine*}} & \multicolumn{2}{c|}{TimeMachine} & \multicolumn{2}{c|}{\textbf{Bi-Mamba*}} & \multicolumn{2}{c|}{Bi-Mamba} & \multicolumn{2}{c|}{iTransformer} & \multicolumn{2}{c|}{TimeMixer} & \multicolumn{2}{c|}{CrossFormer} & \multicolumn{2}{c|}{TimesNet} & \multicolumn{2}{c}{Informer} \\
    \midrule
    Metric & \textbf{MAE} & \textbf{MSE} & \textbf{MAE} & \textbf{MSE} & \textbf{MAE} & \textbf{MSE} & \textbf{MAE} & \textbf{MSE} & \textbf{MAE} & \textbf{MSE} & \textbf{MAE} & \textbf{MSE} & \textbf{MAE} & \textbf{MSE} & \textbf{MAE} & \textbf{MSE} & \textbf{MAE} & \textbf{MSE} \\
    \midrule
    \textbf{ETTh1} & \textbf{0.391 } & \underline{0.389}  & 0.397  & 0.397  & 0.399  & 0.401  & 0.404  & 0.412  & \underline{0.393}  & \textbf{0.386 } & 0.509  & 0.621  & 0.408  & 0.418  & 0.648  & 0.716  & 0.636  & 0.856  \\
    \textbf{ETTh2} & \textbf{0.321 } & \textbf{0.279 } & 0.333  & 0.288  & \underline{0.329}  & \underline{0.286}  & 0.335  & 0.291  & 0.343  & 0.305  & 0.397  & 0.360  & 0.516  & 0.531  & 0.814  & 1.129  & 1.171  & 2.200  \\
    \textbf{ETTm1} & \underline{0.352}  & \underline{0.324}  & 0.361  & 0.344  & \textbf{0.349 } & \textbf{0.318 } & 0.354  & 0.329  & 0.361  & 0.344  & 0.368  & 0.354  & 0.430  & 0.397  & 0.538  & 0.537  & 0.608  & 0.690  \\
    \textbf{ETTm2} & \textbf{0.253 } & \textbf{0.176 } & 0.260  & 0.182  &\underline{0.256}  & \underline{0.177}  & 0.261  & 0.183  & 0.257  & 0.177  & 0.278  & 0.196  & 0.347  & 0.260  & 0.548  & 0.526  & 0.581  & 0.551  \\
    \textbf{Traffic} & \textbf{0.308 } & \textbf{0.451 } & 0.310  & \underline{0.456}  & \underline{0.308}  & 0.460  & 0.310  & 0.462  & 0.321  & 0.468  & 0.420  & 0.699  & 0.309  & 0.583  & 0.337  & 0.693  & 0.438  & 0.832  \\
    \textbf{Electricity} & \textbf{0.262 } & \textbf{0.171 } & 0.268  & \underline{0.177}  & \underline{0.265}  & 0.179  & 0.273  & 0.186  & 0.272  & 0.189  & 0.321  & 0.236  & 0.309  & 0.215  & 0.473  & 0.461  & 0.541  & 0.546  \\
    \bottomrule
    \end{tabular}%
    }
  \caption{Comparative Results Under Missing Points Data. The best results are \textbf{bolded}, and the second-best results are \underline{underlined}.}
  \label{tab:missingpoint}%
\end{table}%

\noindent\textbf{Result} The comparative results are shown in Table \ref{tab:missingpoint}. \textbf{TimeMachine*} and \textbf{Bi-Mamba*} represent the RCL-enhanced versions of TimeMachine and Bi-Mamba, respectively. In the case of missing data points, the backbone models TimeMachine and Bi-Mamba performed worse than other backbone models, such as iTransformer. With the enhancement of RCL, the performance of \textbf{TimeMachine*} and \textbf{Bi-Mamba*} significantly improved, surpassing other models. This demonstrates that RCL can still bring significant improvements to Mamba-based models even under the condition of missing points.

\section{Analysis}

\subsection{Analysis of Time and Memory Overhead}
\label{subsec:Analysis}

Sequence repetition and Repetitive Contrastive Learning introduce additional memory and time overhead. To better understand the implications, we analyze the time and space complexity of the entire training process. The memory overhead for Mamba is determined by the number of blocks, $n_b$, and sequence length, $s_l$, yielding a complexity of $O(s_l n_b)$. During pre-training, only a single Mamba block is utilized, with input sequence lengths $n_ts$ and $s$, resulting in a space complexity of $O((n_t + 1)s)$. Meanwhile, the memory consumption during inference is represented as $O(s n_b)$. Table \ref{tab:timeandspace} details the memory consumption for Mamba training with $n_t = 3$ and $n_b = 4$ layers, illustrating that the peak memory overhead is comparable. As the number of Mamba layers increases, the memory requirement for pre-training remains significantly lower than that of the inference stage.

Due to Mamba's unique computational optimizations, the time complexity of a Mamba block is linear with respect to the sequence length $s_l$, denoted as $O(s_l)$. During pre-training, the sequence length is $n_ts$, whereas during inference, it is $s$. As such, the training time with pre-training is approximately $n_t + 1$ times longer compared to training without pre-training. Table \ref{tab:timeandspace} shows that when $n_t = 3$, the pre-training time consumption is about three times that of inference, which is consistent with our theoretical analysis.

\begin{table}[thbp]
  \centering

    \begin{tabular}{c|ccc|ccc}
    \toprule
          & \multicolumn{3}{c|}{Memory(Unit: MB)} & \multicolumn{3}{c}{Time(Unit: S)} \\
    \midrule
    ETTh1 & Pretrain & Inference & Max Memory & Pretrain & Inference & Total \\
    w/o   & -     & 11733 & 11733 & -     & 1.69  & 1.71 \\
    w     & 13131 & 11470 & 13131 & 5     & 1.62  & 6.54 \\
    \midrule
    Traffic & Pretrain & Inference & Max Memory & Pretrain & Inference & Total \\
    w/o   &     -  &    1602   &    1602   &   -    &   2.67    &  2.68\\
    w     &    1994   &  1298     &   1994    &   6    &   2.54    & 8.54 \\
    \bottomrule
    \end{tabular}%

    \caption{Peak memory consumption and average time overhead. The batch size for the ETTh1 dataset is 2000, while for the Traffic dataset it is 100.}
  \label{tab:timeandspace}%
\end{table}%

\subsection{Analysis of Enhanced Selectivity}\label{subsec:AnalysisSelectivity}

\noindent\textbf{Through Focus Ratio and Memory Entropy}\\
We compared the Focus Ratio and Memory Entropy of the Mamba block when modeling time series with and without RCL, and the results are shown in Table 7. The statistical significance test and correlation analysis of Focus Ratio (FR) and Memory Entropy (ME) are presented in \ref{ssca}.  It can be observed that after applying RCL, the Focus Ratio significantly improves, indicating more pronounced processes of significant memory and significant forgetting. This suggests that the model becomes more focused on key information and more decisive in ignoring noisy information. Similarly, RCL also leads to a notable increase in Memory Entropy, demonstrating that the Mamba block's memory patterns for time-step information become more complex and diverse. This enables the model to better capture essential aspects of the sequence with greater selectivity.

\begin{figure}[ht]
    \centering
    \begin{minipage}[t]{7cm}
        \centering
        \includegraphics[width=7cm]{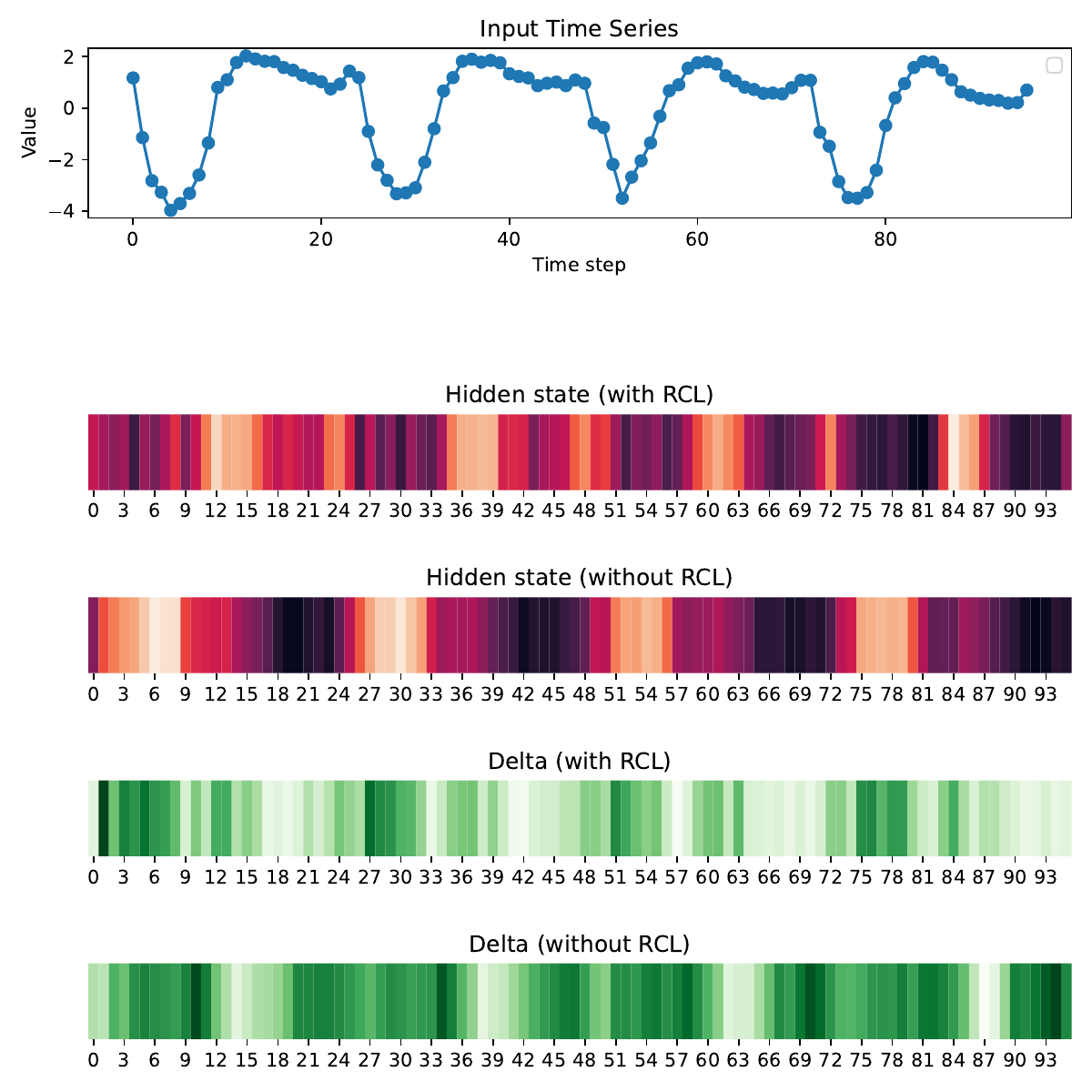}
        \caption{Hidden state and $\Delta$ corresponding to the input time series.}
        \label{fig:virtual}
    \end{minipage}
    \hfill
    \begin{minipage}[t]{7cm}
        \centering
        \includegraphics[width=7cm]{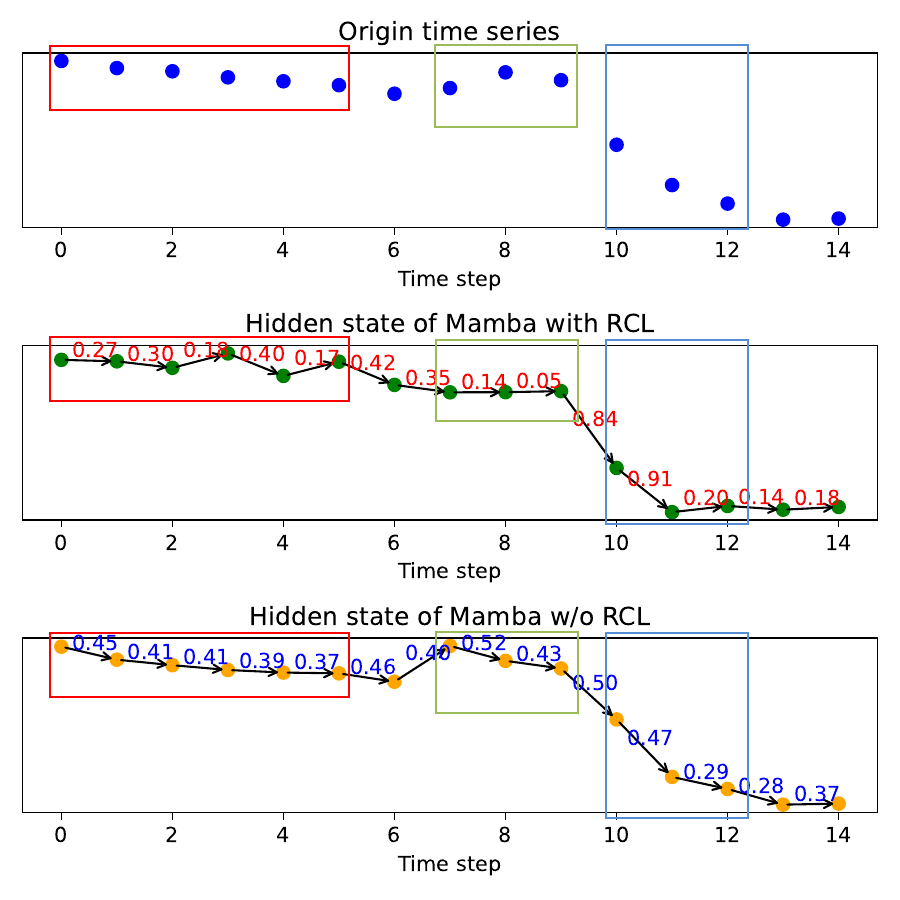}
        \caption{Memory and ignoring in Mamba models with and without RCL.}
        \label{fig:tsmemory}
    \end{minipage}
\end{figure}

\noindent\textbf{Through Visualization of the Hidden state and Delta}\\
We demonstrate that our proposed RCL effectively enhances the time step selection capability of the Mamba block by visualizing the Hidden state and Delta corresponding to the input time series of the Mamba block. The visualization results are shown in Figure \ref{fig:virtual}. According to the principles of SSM, the Hidden state can be represented in a form similar to a recurrent neural network:

\begin{equation}
    \textbf{H}_{t+1} = \overline{\textbf{A}} \textbf{H}_{t} + \overline{\textbf{B}} \textbf{X}_{t+1}
\end{equation}

The matrix $\textbf{A}$ determines how historical temporal information is retained. In the Mamba block, the matrix $\overline{\textbf{A}}$ is determined by a fixed matrix $\textbf{A}$ and $\Delta$, where $\textbf{A}$ influences part of the historical information selection, and $\Delta$ influences another part. The visualization results indicate that without initializing with RCL parameters, the Hidden state is almost directly proportional to the input, and $\Delta$ is similarly proportional to the input. This suggests that directly training the Mamba block does not effectively retain historical information; the matrix $\textbf{A}$ nearly forgets all historical information, retaining only the current information as the hidden state.

In contrast, when training with initialized parameters, the Hidden state exhibits more complex representations, and $\Delta$ shows a more intricate temporal pattern. This indicates that the model learns complex inter-dependencies between time steps. The matrix $\textbf{A}$ learned by RCL demonstrates different memory and ignoring patterns for historical information across various time steps. It retains more of the input at critical time steps while preserving more historical information at non-critical time steps, thereby significantly enhancing the Mamba block's ability to select relevant information from time series data.



\textbf{Through Visualization of Memory and Ignoring Processes}
We visualized the evolution of the memory score for a sequence when using RCL and when not using RCL, where the definition of the memory score is provided in Section \ref{measure}.The visualization results are presented in Figure \ref{fig:tsmemory}, where the numbers on the arrows indicate the memory weights for the previous time step. From the perspective of recurrent neural networks, it is evident that without using RCL for parameter initialization, the Mamba block maintains historical memory weights between 30\% and 50\% across all time steps, resulting in a hidden state that closely resembles the original time series. In contrast, the Mamba block with RCL exhibits a richer memory pattern, demonstrating significant noise resistance and strong memory retention for critical time steps.

In region (a) of Figure \ref{fig:tsmemory}, the original time series is monotonically decreasing. Here, the hidden state of Mamba w/o RCL is almost identical to the original time series, while Mamba with RCL maintains the overall downward trend but differentiates the spatial representation of each time step, resulting in more pronounced changes in the hidden state. In region (b), a brief noise appears amidst the overall decline. Mamba w/o RCL is noticeably affected by this noise, whereas Mamba with RCL overcomes the noise interference by leveraging high historical memory weights. In region (c), the original time series experiences a significant drop. Mamba with RCL accurately identifies the critical points of this abrupt change, largely ignoring historical information to prominently incorporate the crucial time step information into the hidden state.

\section{ Visualization of Comparative Effects}\label{apdx_vir}
\subsection{Visualization of Embedding in Augmentation Sequence}

\begin{figure}[htbp]
    \centering
    \begin{minipage}{7cm}
        \centering
        \includegraphics[width=7cm]{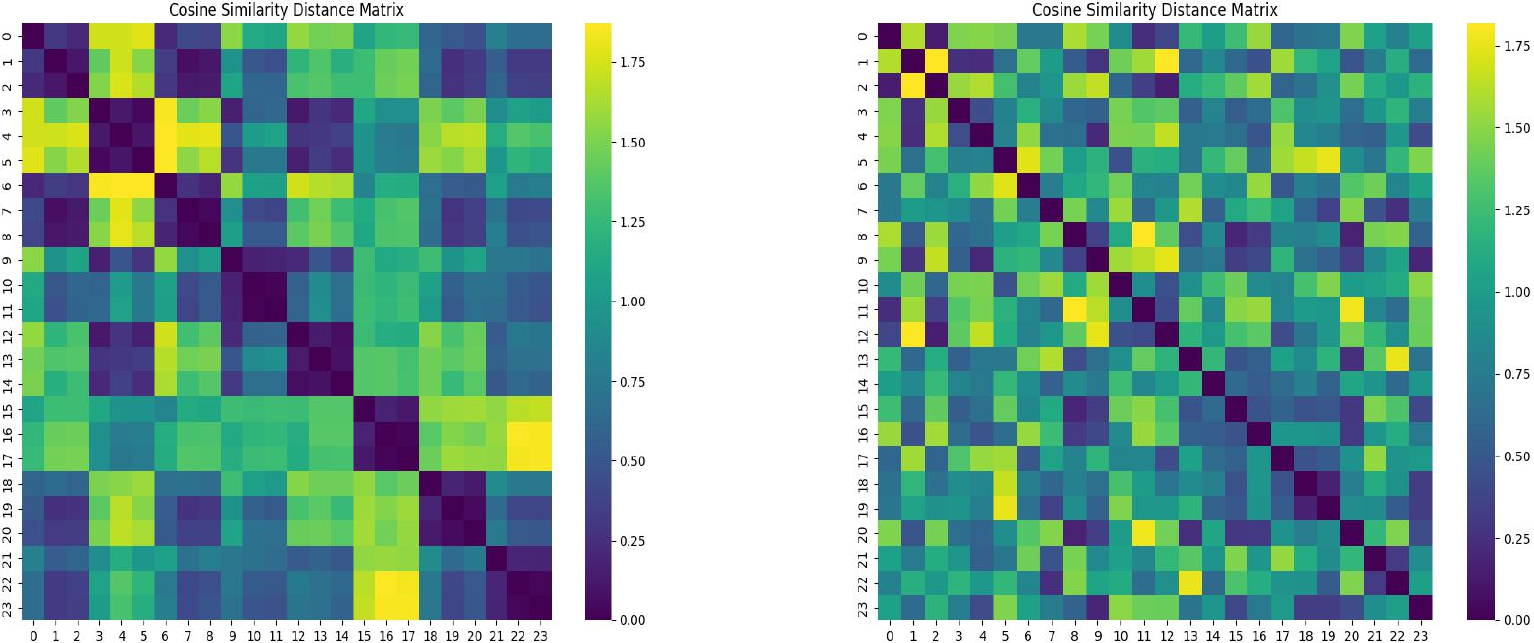}
        \caption{Embedding with contrastive pre-training result}
        \label{fig:3}
    \end{minipage}
    \hspace{0.5cm}
    \begin{minipage}{7cm}
        \centering
        \includegraphics[width=7cm]{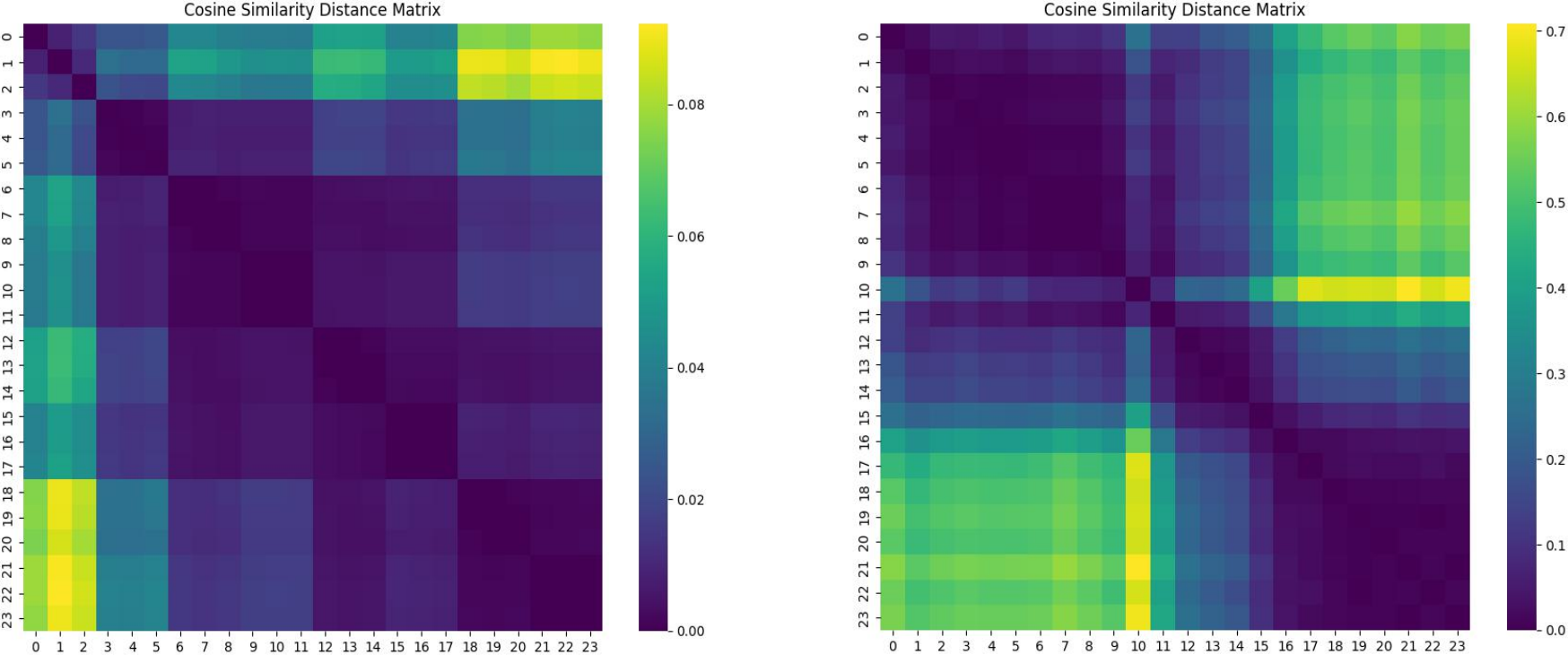}
        \caption{Embedding without pre-training result}
        \label{fig:4}
    \end{minipage}
\end{figure}

To visually demonstrate the impact of our contrastive learning methods, we plotted the cosine similarity values\citep{inproceedings} between embedding vectors of the same input sequence from the ETTm1 dataset using a heatmap \citep{JSSv016c03}. This comparison involves identical Mamba blocks—one trained without contrastive pre-training and the other with it. The resulting variations in distribution highlight the influence of our pre-training objectives, which enhance the model’s ability to selectively focus on relevant features. The images illustrate the differences in the embedding space (Figure \ref{fig:4}) and the refined distribution achieved through contrastive learning (Figure \ref{fig:3}).

It is evident that the Mamba model without RCL struggles to effectively distinguish between irrelevant noise and valid time steps, and it fails to make effective selections within the time series. Additionally, the original Mamba model cannot adequately separate different time steps, maintaining high correlation, which indicates that new time step information fails to be effectively encoded and merely perturbs the coding. In contrast, Mamba with RCL effectively differentiates between valid time steps and filters out noise, mitigating the effects of long sequences and introducing more valid information, thereby improving the modeling of the entire sequence.

\subsection{Visualization of clustering of Positive and Negative Cases}

\begin{figure}[htbp]
    \centering
    \begin{minipage}{10cm}
        \centering
        \includegraphics[width=10cm]{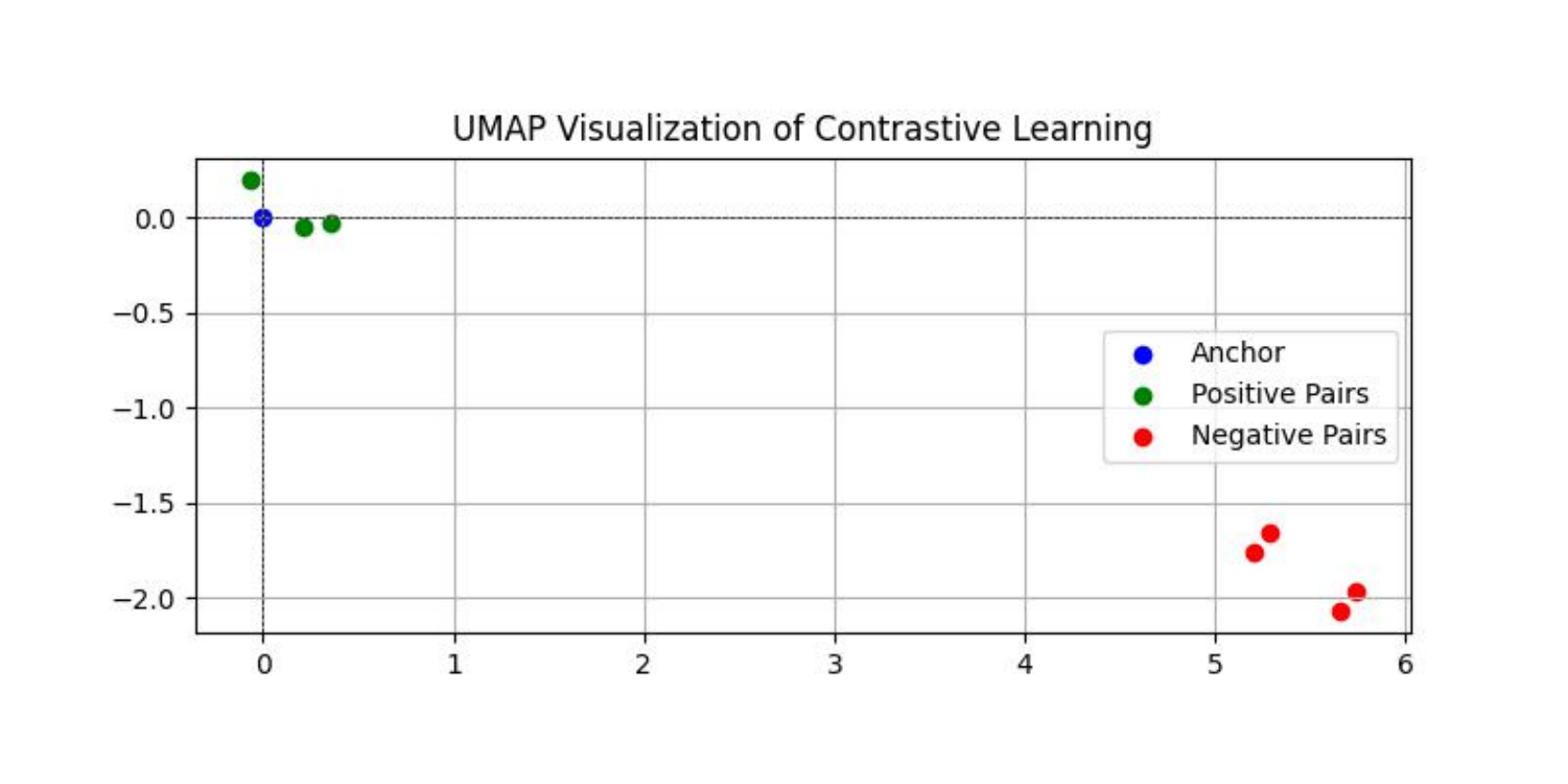}
        \caption{UMAP reduction results. Anchor points are randomly selected, and all other points are related to the anchor.}
        \label{fig:5}
    \end{minipage}
\end{figure}

We also visualized the detailed distribution of vectors using the UMAP technique for dimensionality reduction, where the original dimensionality of the embedding vectors is 32. UMAP is based on a theoretical framework rooted in Riemannian geometry and algebraic topology, resulting in a scalable and practical algorithm suitable for contrastive learning data \citep{mcinnes2020umapuniformmanifoldapproximation}. In the visualizations (Figure \ref{fig:5}), we randomly selected embedding vectors from input sequences and plotted the corresponding vectors for both positive and negative pairs in our method.

The clustering results demonstrate that the model can effectively distinguish between positive and negative examples, with positive examples clustering near the anchor and negative examples retreating farther away. The significance of this distinction is evident in the clustering results, indicating that our method can better recognize valid and invalid time steps, and possesses stronger differentiation and selection capabilities.

\subsection{Case Study}
\begin{figure}
    \centering
    \includegraphics[width=\linewidth]{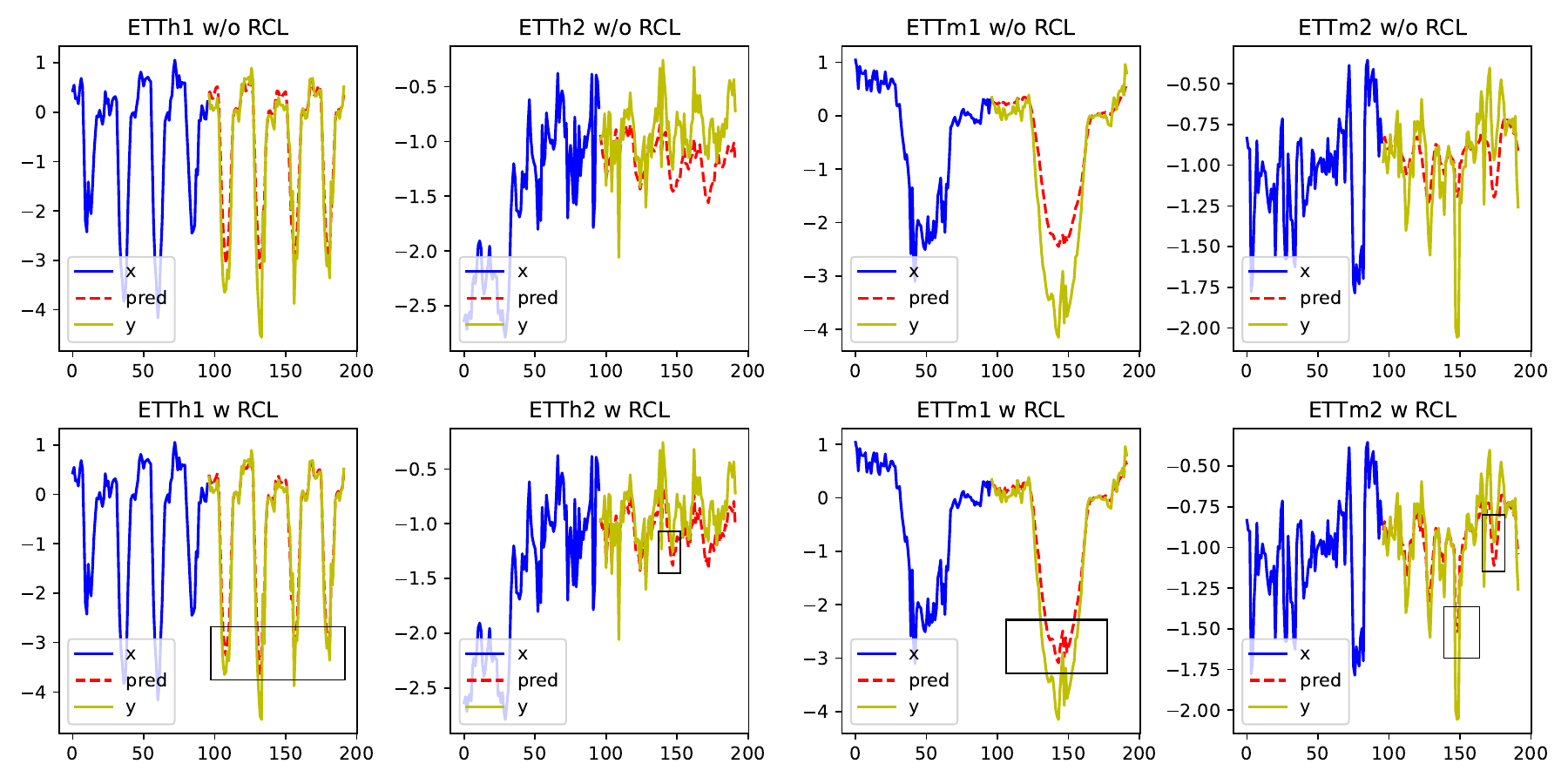}
    \caption{Visualization of prediction results with and without RCL across various datasets.}
    \label{fig:casestudy}
\end{figure}
We conducted a case study on multiple datasets for both with RCL and without RCL scenarios, and the visualization results are shown in Fig, \ref{fig:casestudy}. The backbone model selected is Timemachine. It can be observed that after incorporating RCL, the model is able to make more accurate predictions, with the predicted curve (red) closely aligning with the actual curve (yellow). More importantly, with RCL, the backbone model can make more accurate predictions and adapt more quickly to extreme cases. As highlighted in Fig, \ref{fig:casestudy}, the predicted values with RCL are significantly closer to the true values in extreme scenarios. This demonstrates that RCL is a highly effective method, capable of adapting to extreme changes in time series patterns through enhanced memory and forgetting mechanisms for extreme cases.

\section{Related Work}

\subsection{Models in Deep Time Series Forecasting}

Extensive research has been conducted to address time series forecasting problems, primarily focused on proposing new models that improve prediction accuracy. These models can be categorized into five primary groups: Transformer-based, RNN-based\cite{10.1162/neco.1997.9.8.1735}, CNN-based, MLP-based, Mamba-based and spatio-temporal models which use extra graph data. While emphasizing different aspects, these approaches aim to address key challenges of time series tasks.

Some MLP-based models, such as N-Beats\cite{oreshkin2020nbeatsneuralbasisexpansion} and N-HiTs\cite{challu2022nhitsneuralhierarchicalinterpolation}, utilize basis approximation and residual connections. TimesNet \cite{wu2023timesnettemporal2dvariationmodeling}, a CNN-based model, employs periodical segmentations in the frequency and time domains, extracting inter-period and intra-period patterns. TimeMixer \cite{wang2024timemixerdecomposablemultiscalemixing}, built solely with MLP and pooling layers, excels by decomposing and mixing multi-scale data.

Transformer-based models like LogTrans\cite{li2020enhancinglocalitybreakingmemory}, Informer\cite{zhou2021informerefficienttransformerlong}, Autoformer\cite{wu2022autoformerdecompositiontransformersautocorrelation}, TCGPN\cite{yan2024tcgpn}, DPA-STIFormer\cite{yan2024double} and FEDformer\cite{zhou2022fedformerfrequencyenhanceddecomposed} enhance adaptability using sparse attention and decomposition techniques. PatchTST\cite{nie2023timeseriesworth64} segments time series into patches for denoising, while iTransformer\cite{liu2024itransformerinvertedtransformerseffective} redefines time embeddings. 

Mamba-based models like TimeMachine\cite{ahamed2024timemachinetimeseriesworth}, HIGSTM\cite{yan2025hierarchical} combine channel-mixing and independence to enhance content selection, while S-Mamba\cite{wang2024mambaeffectivetimeseries} achieves transformer-like generalization with reduced computational resources.

Spatio-temporal models like STGNN \cite{ALIGHTWEIGHT}decompose time series into seasonal and trend components using downsampling and patched sampling, while SAMBA\cite{MAMBAFINANCIAL} employs bidirectional Mamba blocks and adaptive graph convolution to model long-term dependencies and spatial stock feature interactions with near-linear complexity, and DSTGCN\cite{10770607} integrates graph convolution with spatial attention to efficiently capture spatial relationships.





\subsection{Contrastive Learning}

Most contrastive self-supervised learning methods have been applied in vision \cite{jaiswal2021surveycontrastiveselfsupervisedlearning} and multimodal learning \cite{manzoor2024multimodalityrepresentationlearningsurvey}, leveraging high-level attributes that are easily distinguishable and less affected by noise. For example, images remain interpretable despite perturbations like color changes or geometric transformations, while multimodal methods enhance contrast by using cross-modality correlations, such as visual-textual pairing.

In contrast, applying contrastive learning to unimodal sequential data is less common and often requires tailored features. For instance, CodeRetriever \cite{li2022coderetrieverunimodalbimodalcontrastive} employs similarity contrastive loss to capture nuances in code sequences. ECP-Mamba\cite{ECPMAMBA} integrate multi-scale self-supervised contrastive learning with a state space model (SSM) backbone in image modality. Sequential recommendation \cite{XIE2024112257} and text summarization \cite{xu2022sequencelevelcontrastivelearning} rely on specialized sequence representations and training techniques. 

In time series, contrastive pre-training has improved representation learning. TS2Vec \cite{yue2022ts2vecuniversalrepresentationtime} introduced a universal framework using context view augmentation and hierarchical contrastive learning. TF-C \cite{zhang2022selfsupervisedcontrastivepretrainingtime} aligned time-based and frequency-based representations for better performance. InfoTS applied information theory to prioritize diverse and high-fidelity representations, while SoftCLT \cite{softcls} captured inter-sample and intra-temporal relationships through soft assignments.

These methods excel in representation learning for classification but are less effective for forecasting. Our approach pre-trains mamba models to capture recurrent noise patterns, enabling the direct application of pre-trained parameters to forecasting tasks, representing a novel and significant improvement over existing methods.

\section{Conclusion}
In this paper, we introduce Repetitive Contrastive Learning (RCL), a novel training paradigm designed to enhance the selective capabilities of Mamba blocks and enable the transfer of these parameters to various Mamba-based backbone models, improving their performance. RCL combines sequence repetition with intra-sequence and inter-sequence contrastive learning, strengthening Mamba blocks' ability to retain critical information and filter out noise. Through extensive experiments, we demonstrate RCL's effectiveness across multiple Mamba-based backbone models and diverse temporal prediction tasks, significantly enhancing their temporal modeling and forecasting capabilities. From theoretical, qualitative, and quantitative perspectives, we validate the enhanced selective performance achieved by RCL and confirm that it adds no extra memory overhead. In addition, we have validated the generalization capability of RCL across other types of models. Through experiments, we have demonstrated that RCL is significantly effective on step-by-step models like RNNs, and we have analyzed why it is not well-suited for non-step-by-step models such as Transformers.

\section*{Acknowledgments}
This work is supported by the National Natural Science Foundation of China (Grant No. 62276008, 62250037, and 62076010), and partially supported by the National Key R\&D of China (Grant \#2022YFF0800601).

\bibliographystyle{elsarticle-harv} 
\bibliography{example_paper}




\clearpage
\appendix
\section{Reproducibility Statement}
We provide simplified code available at this Github link\footnote{https://github.com/xiaxiaoguang/PretrainMamba}. You can use this code to reproduce our results by referring to the parameters outlined in the paper.

\section{Algorithm Workflow}

In this section, we introduce the workflow of our algorithm, which consists of two stages. First, we perform Repetitive Contrastive Learning on a single Mamba block to obtain the trained parameters$\mathbf{A}$, $\mathbf{B}$, and $\mathbf{C}$. The algorithm is illustrated in Algorithm \ref{alg:repetitive_contrastive_learning}.
\begin{algorithm}[ht]
\caption{Repetitive Contrastive Learning}
\label{alg:repetitive_contrastive_learning}
\begin{algorithmic}[1]
\Require Input $\mathbf{X}$: $(B, T, F)$
\Ensure Outputs $\mathbf{A}$, $\mathbf{B}$, $\mathbf{C}$
    \State $\mathbf{X}_{rep}$: $(B, T, n_t, F)$  $\leftarrow$ repeating each time step $n_t$ times
    \For{$j = 1$ to $n_t-1$}
        \State Sample $Noise_{j} \sim \mathcal{N}(0, \sigma_j^2)$: $(B, T, F)$
    \EndFor
    \State $Noise_0 = 0$
    \State $Noise$: $(B, T, n_t, F)$  $\leftarrow$ Stack $ [Noise_{0}, \dots, Noise_{n_t-1}]$
    \State $\mathbf{X}_{aug}$: $(B, T, n_t, F)$   $\leftarrow$ $\mathbf{X}_{rep} + Noise$
    \State $\mathbf{H}$:$(B, T,  D)$ $\leftarrow$  $\text{MambaBlock}(\mathbf{X})$
    \State $\mathbf{H}_{aug} $: $(B, T, n_t, D)$ $\leftarrow$  $\text{MambaBlock}(X_{aug})$
    \State $\mathcal{L}_{\text{Intra}}$: Compute  Intra-sequence contrast
    \State $\mathcal{L}_{\text{Inter}}$: Compute Inter-sequence contrast
\State \Return $\mathbf{A}$, $\mathbf{B}$, $\mathbf{C}$
\end{algorithmic}
\end{algorithm}

In the second stage, an arbitrary Mamba-based backbone model is selected, and the initialization parameters of its Mamba blocks are replaced with $\mathbf{A}$, $\mathbf{B}$, $\mathbf{C}$. The algorithm is illustrated in Algorithm \ref{alg:replace_and_train}.

\begin{algorithm}
\caption{Replace and Train Backbone Model}
\label{alg:replace_and_train}
\begin{algorithmic}[2]
\Require Input $\mathbf{X}$: $(B, T, F)$, model $\mathbb{M}$, parameters $\mathbf{A}$, $\mathbf{B}$, $\mathbf{C}$
\Ensure Trained model $\mathbb{M}'_{aug}$
    \State $\mathbb{M}_{aug}$: Replace Mamba Block`s parameters in $\mathbb{M}$ with $\mathbf{A}$, $\mathbf{B}$, $\mathbf{C}$
    \State $\mathbb{M}'_{aug}$: Train $\mathbb{M}_{aug}$ using $\mathbf{X}$
    \State \Return $M'_{aug}$
\end{algorithmic}
\end{algorithm}

\section{Theoretical Analysis of Repetitive Contrastive Learning}\label{apdx_theo}
We conducted a theoretical analysis of RCL with a single repetition. For S4 SSM, where A and B are fixed, for any anchor $ h_{t} $, the positive example is the time step after repetition $ h_t^{+} = Ah_{t}+B(x_{t}+\sigma) $, and the negative example is the next time step $ h_{t}^{-} = Ah_{t}^{+}+Bx_{t+1} = A^2h_{t} + AB(x_t+\sigma) + Bx_{t+1} $. We measure relevance using cosine similarity, assuming all vectors are normalized, so cosine similarity simplifies to $ sim(a,b) = a \cdot b $.

$$\text{Sim}_{\text{pos}} = sim(h_t, h_t^{+}) = Ah_t^2 + B(x_t+\sigma)h_t
$$
$$\text{Sim}_{\text{neg}} = sim(h_t, h_t^{-}) = A^2h_t^2 + AB(x_t+\sigma)h_t + Bx_{t+1}h_{t}$$

Assuming a temperature coefficient of 1, the contrastive loss can be written as:

$$\text{Loss} = \log(1 + \exp(\text{Sim}_{\text{neg}} - \text{Sim}_{\text{pos}}))
$$

Minimizing the InfoNCE loss can be interpreted as maximizing the lower bound of mutual information between the anchor and the positive example\citep{oord2019representationlearningcontrastivepredictive,wu2020mutualinformationcontrastivelearning}:

$$I(h_t, h_t^{+}) \geq -\text{Loss} = -\log(1 + \exp(\text{Sim}_{\text{neg}} - \text{Sim}_{\text{pos}}))$$

This means making the representations of the noise-free and the next noisy time step similar. Essentially, Mamba aims to remove the interference of noisy time steps and maintain the hidden state unchanged.

Maximizing the lower bound of mutual information is equivalent to minimizing:
$$\text{Sim}_{\text{neg}} - \text{Sim}_{\text{pos}} = (A^2 - A)h_t^{2} + (AB - B)(x_t + \sigma)h_t + Bx_{t+1}h_t$$
By taking derivatives with respect to A and B, we can solve for the optimal $ A^* $ and $ B^* $:
$$A^* = \frac{h_t^2 - B(x_t + \sigma)h_t}{2h_t^2}
$$
$$B^* = \frac{h_t(2x_{t+1} - x_t - \sigma)}{(x_t + \sigma)^2}
$$
The optimal lower bound is:
$$-\log\left(1 + \exp\left(\frac{x_{t+1}^2 - x_{t+1}(x_t + \sigma)}{(x_t + \sigma)^2}h_t^2\right)\right)
$$
As we gradually increase $ \sigma $, the lower bound of mutual information continues to improve, indicating that S4's resistance to noise is enhanced. This is reflected in $ h_t^{+} = Ah_{t} + B(x_{t} + \sigma) $, where A tends to 1 and B tends to 0, emphasizing the selection of historical information while ignoring noisy time steps.

Furthermore, we analyze the state transition in Mamba. In Mamba, matrices A and B can be approximately considered as linearly transformed from the current time step:
$$A_{+} = W_A(x_t + \sigma), \quad A_{-} = W_Ax_{t+1}, \quad B_{+} = W_B(x_t + \sigma), \quad B_{-} = W_Bx_{t+1}
$$
Substituting into the loss function and taking derivatives with respect to $ W_A $ and $ W_B $, the optimal lower bound of mutual information is:
$$-\log\left(1 + \exp\left(\frac{h_t^2x_{t+1}^{4}}{(x_t + \sigma)^3} + \frac{h_t^2}{x_t + \sigma}\right)\right)
$$
An increase in noise intensity enhances the lower bound of mutual information, and the optimal lower bound is more sensitive to noise.

\section{Additional Experimental Results}\label{apdx_addre}

\subsection{Comprehensive Results of Comparison with Temporal Model}\label{full_compare}
Table \ref{tab:tcomp} provides the detailed comparative results across the four prediction horizons, where we achieve significant improvements and attain state-of-the-art performance across multiple prediction lengths.
\begin{table*}[htbp]
  \centering

  \resizebox{\textwidth}{!}{

\begin{tabular}{c|c|cc|cc|cc|cc|cc|cc|cc|cc|cc|cc|cc|cc|cc}
    \toprule
    \multicolumn{2}{c|}{Model} & \multicolumn{2}{c|}{\textbf{TimeMachine*}} & \multicolumn{2}{c|}{TimeMachine} & \multicolumn{2}{c|}{\textbf{Bi-Mamba*}} & \multicolumn{2}{c|}{Bi-Mamba} & \multicolumn{2}{c|}{iTransformer} & \multicolumn{2}{c|}{TimeMixer} & \multicolumn{2}{c|}{CrossFormer} & \multicolumn{2}{c|}{PatchTST} & \multicolumn{2}{c|}{TimesNet} & \multicolumn{2}{c|}{FEDFormer} & \multicolumn{2}{c|}{Informer} & \multicolumn{2}{c|}{N-HiTS} & \multicolumn{2}{c}{ N-BEATS} \\
    \midrule
    \multicolumn{2}{c|}{Metric} & MAE   & MSE   & MAE   & MSE   & MAE   & MSE   & MAE   & MSE   & MAE   & MSE   & MAE   & MSE   & MAE   & MSE   & MAE   & MSE   & MAE   & MSE   & MAE   & MSE   & MAE   & MSE   & MAE   & MSE   & MAE   & MSE \\
    \midrule
    \multirow{4}[2]{*}{ETTh1} & 96    & \textbf{0.387 } & \underline{0.379}  & 0.391  & 0.383  & \underline{0.389}  & \underline{0.379}  & 0.395  & 0.381  & 0.405  & 0.386  & 0.400  & \textbf{0.375 } & 0.448  & 0.423  & 0.419  & 0.414  & 0.402  & 0.384  & 0.419  & 0.376  & 0.713  & 0.865  & 0.397  & 0.394  & 0.415  & 0.406  \\
          & 192   & \textbf{0.420 } & 0.440  & 0.423  & 0.440  & \underline{0.421}  & \textbf{0.425 } & 0.428  & \underline{0.427}  & 0.436  & 0.441  & 0.421  & 0.429  & 0.474  & 0.471  & 0.445  & 0.460  & 0.429  & 0.436  & 0.448  & 0.420  & 0.792  & 1.008  & 0.434  & 0.478  & 0.514  & 0.535  \\
          & 336   & \textbf{0.442 } & \underline{0.482}  & \underline{0.446 } & 0.490  & 0.456  & \textbf{0.481 } & 0.459  & 0.484  & 0.458  & 0.487  & 0.458  & 0.484  & 0.546  & 0.570  & 0.466  & 0.501  & 0.469  & 0.491  & 0.465  & 0.459  & 0.809  & 1.107  & 0.489  & 0.508  & 0.499  & 0.495  \\
          & 720   & \textbf{0.466 } & \textbf{0.488 } & \underline{0.470}  & \underline{0.496}  & 0.496  & \underline{0.496}  & 0.496  & 0.516  & 0.491  & 0.503  & 0.482  & 0.498  & 0.621  & 0.653  & 0.488  & 0.500  & 0.500  & 0.521  & 0.507  & 0.506  & 0.865  & 1.181  & 0.499  & 0.519  & 0.523  & 0.523  \\
    \midrule
    \multirow{4}[2]{*}{ETTh2} & 96    & \textbf{0.330 } & \textbf{0.282 } & \underline{0.334}  & 0.291  & 0.347  & 0.300  & 0.349  & 0.307  & 0.349  & 0.297  & 0.341  & \underline{0.289}  & 0.584  & 0.745  & 0.348  & 0.302  & 0.374  & 0.340  & 0.397  & 0.358  & 1.525  & 3.755  & 0.346  & 0.303  & 0.331  & 0.233  \\
          & 192   & \textbf{0.382 } & \textbf{0.355 } &\underline{ 0.385}  & \underline{0.369 } & 0.394  & 0.373  & 0.398  & 0.377  & 0.400  & 0.380  & 0.392  & 0.372  & 0.656  & 0.877  & 0.400  & 0.388  & 0.414  & 0.402  & 0.439  & 0.429  & 1.931  & 5.602  & 0.417  & 0.396  & 0.432  & 0.372  \\
          & 336   & \underline{0.420 } &\underline{ 0.412 } & 0.428  & 0.421  & 0.429  & 0.434  & 0.434  & 0.435  & 0.432  & 0.428  & \textbf{0.414 } & \textbf{0.386 } & 0.731  & 1.043  & 0.433  & 0.426  & 0.452  & 0.452  & 0.487  & 0.496  & 1.835  & 4.721  & 0.514  & 0.468  & 0.507  & 0.479  \\
          & 720   & \textbf{0.430 } & \textbf{0.412 } & 0.439  & 0.424  & 0.602  & 0.731  & 0.597  & 0.715  & 0.445  & 0.427  & \underline{0.434}  & \underline{0.412}  & 0.763  & 1.104  & 0.446  & 0.431  & 0.468  & 0.462  & 0.474  & 0.463  & 1.625  & 3.647  & 0.514  & 0.518  & 0.616  & 0.560  \\
    \midrule
    \multirow{4}[2]{*}{ETTm1} & 96    & \textbf{0.346 } & \textbf{0.318 } & 0.361  & 0.334  & 0.358  & 0.332  & 0.364  & 0.332  & 0.368  & 0.334  & \underline{0.357}  &\underline{0.320 } & 0.426  & 0.404  & 0.367  & 0.329  & 0.375  & 0.338  & 0.419  & 0.379  & 0.571  & 0.672  & 0.371  & 0.352  & 0.378  & 0.364  \\
          & 192   & \textbf{0.377 } &\underline{ 0.375}  & \underline{0.379}  & 0.379  & 0.384  & 0.369  & 0.389  & 0.378  & 0.391  & 0.377  & 0.381  & \textbf{0.361 } & 0.451  & 0.450  & 0.385  & 0.367  & 0.387  & 0.374  & 0.441  & 0.426  & 0.669  & 0.795  & 0.396  & 0.389  & 0.385  & 0.381  \\
          & 336   & \textbf{0.387 } & \underline{0.396}  & \underline{0.394}  & 0.401  & 0.407  & 0.404  & 0.412  & 0.405  & 0.420  & 0.426  & 0.404  & \textbf{0.390 } & 0.515  & 0.532  & 0.410  & 0.399  & 0.411  & 0.410  & 0.459  & 0.445  & 0.871  & 1.212  & 0.393  & 0.413  & 0.401  & 0.399  \\
          & 720   & \textbf{0.429 } & \underline{0.455}  &\underline{ 0.431}  & 0.467  & 0.441  & 0.458  & 0.452  & 0.466  & 0.459  & 0.491  & 0.441  & \textbf{0.454 } & 0.589  & 0.666  & 0.439  & 0.454  & 0.450  & 0.478  & 0.490  & 0.543  & 0.823  & 1.166  & 0.502  & 0.487  & 0.441  & 0.529  \\
    \midrule
    \multirow{4}[2]{*}{ETTm2} & 96    & \textbf{0.251 } & \textbf{0.173 } & 0.253  & \underline{0.175}  & 0.271  & 0.186  & 0.270  & 0.188  & 0.264  & 0.180  & 0.258  &\underline{0.175}  & 0.366  & 0.287  & 0.259  & 0.175  & 0.267  & 0.187  & 0.287  & 0.203  & 0.453  & 0.365  & 0.255 & 0.176 & 0.263 & 0.184 \\
          & 192   & \textbf{0.293 } & \underline{0.238}  & \underline{0.294}  & 0.238  & 0.313  & 0.254  & 0.315  & 0.257  & 0.309  & 0.250  & 0.299  & \textbf{0.237 } & 0.492  & 0.414  & 0.302  & 0.241  & 0.309  & 0.249  & 0.328  & 0.269  & 0.563  & 0.533  & 0.305 & 0.245 & 0.337 & 0.273 \\
          & 336   & \textbf{0.333 } & \underline{0.299}  & \underline{0.337}  & 0.307  & 0.364  & 0.316  & 0.387  & 0.392  & 0.348  & 0.311  & 0.340  & \textbf{0.298 } & 0.542  & 0.597  & 0.343  & 0.305  & 0.351  & 0.321  & 0.366  & 0.325  & 0.887  & 1.363  & 0.346 & 0.295 & 0.355 & 0.309 \\
          & 720   & \textbf{0.392 } &\underline{ 0.402}  & \underline{0.394}  & 0.407  & 0.413  & 0.404  & 0.430  & 0.429  & 0.407  & 0.412  & 0.396  & \textbf{0.391 } & 1.042  & 1.730  & 0.400  & 0.402  & 0.403  & 0.408  & 0.415  & 0.421  & 1.338  & 3.379  & 0.413 & 0.401 & 0.425 & 0.411 \\
    \midrule
    \multirow{4}[2]{*}{Traffic} & 96    & 0.299  & 0.484  & 0.306  & 0.498  & \underline{0.276}  & 0.579  & 0.279  & 0.587  & \textbf{0.268 } & \textbf{0.395 } & 0.285  & \underline{0.462}  & 0.290  & 0.522  & 0.359  & 0.544  & 0.321  & 0.593  & 0.366  & 0.587  & 0.368  & 0.274  & 0.282 & 0.402 & 0.282 & 0.398 \\
          & 192   & \textbf{0.273 } & \textbf{0.412 } & \underline{0.274}  & \underline{0.417}  & 0.308  & 0.625  & 0.306  & 0.630  & 0.276  & \underline{0.417}  & 0.296  & 0.473  & 0.293  & 0.530  & 0.354  & 0.540  & 0.336  & 0.617  & 0.373  & 0.604  & 0.386  & 0.296  & 0.297 & 0.42  & 0.293 & 0.409 \\
          & 336   & \textbf{0.279 } & \textbf{0.429 } & \underline{0.281}  & \underline{0.433}  & 0.311  & 0.666  & 0.307  & 0.659  & 0.283  & \underline{0.433}  & 0.296  & 0.498  & 0.305  & 0.558  & 0.358  & 0.551  & 0.336  & 0.629  & 0.383  & 0.621  & 0.394  & 0.300  & 0.313 & 0.448 & 0.318 & 0.449 \\
          & 720   & \textbf{0.298 } & \textbf{0.459 } & \underline{0.300}  & \underline{0.467}  & 0.336  & 0.689  & 0.338  & 0.702  & 0.302  & \underline{0.467}  & 0.313  & 0.506  & 0.328  & 0.589  & 0.375  & 0.586  & 0.350  & 0.640  & 0.382  & 0.626  & 0.439  & 0.373  & 0.353 & 0.539 & 0.391 & 0.589 \\
    \midrule
    \multirow{4}[2]{*}{Electricity} & 96    & \underline{0.259}  & 0.183  & 0.261  & 0.187  & 0.261  & \underline{0.182}  & 0.263  & 0.185  & \textbf{0.240 } & \textbf{0.148 } & 0.247  & \underline{0.153}  & 0.314  & 0.219  & 0.285  & 0.195  & 0.272  & 0.168  & 0.308  & 0.193  & 0.391  & 0.719  & 0.285  & 0.182  & 0.235  & 0.173  \\
          & 192   & \textbf{0.246 } & \textbf{0.152 } & \underline{0.250}  & \underline{0.158}  & 0.270  & 0.188  & 0.272  & 0.191  & 0.253  & 0.162  & 0.256  & 0.166  & 0.322  & 0.231  & 0.289  & 0.199  & 0.289  & 0.184  & 0.315  & 0.201  & 0.379  & 0.696  & 0.300  & 0.228  & 0.287  & 0.185  \\
          & 336   & \textbf{0.261 } & \textbf{0.169 } & \underline{0.268}  & \underline{0.172}  & 0.283  & 0.200  & 0.290  & 0.212  & 0.269  & 0.178  & 0.277  & 0.185  & 0.337  & 0.246  & 0.305  & 0.215  & 0.300  & 0.198  & 0.329  & 0.214  & 0.420  & 0.777  & 0.354  & 0.242  & 0.355  & 0.257  \\
          & 720   & \textbf{0.295 } & \textbf{0.201 } & \underline{0.298}  & \underline{0.207}  & 0.317  & 0.255  & 0.323  & 0.259  & 0.317  & 0.225  & 0.310  & 0.225 & 0.363  & 0.280  & 0.337  & 0.256  & 0.320  & 0.220  & 0.355  & 0.246  & 0.472  & 0.864  & 0.377  & 0.331  & 0.438  & 0.369  \\
    \bottomrule
    \end{tabular}%

}
  \caption{Comparison results with temporal model. Bolded numbers indicate optimal results and underscores indicate sub-optimal results.}
     \label{tab:tcomp}
\end{table*}

\subsection{Statistical Significance and Correlation Analysis of Focus Ratio and Memory Entropy}\label{ssca}
\begin{table}[h!]
\centering
\resizebox{0.5\linewidth}{!}{
\label{tab:correlation}
\begin{tabular}{c|cc}
\toprule
& \textbf{Pearson Correlation \(r\)} & \textbf{\(p\)-value} \\
\midrule
MSE \& FR & -0.5096 & \(3.1278 \times 10^{-21}\) \\
MSE \& ME & -0.8000 & \(4.4317 \times 10^{-68}\) \\
\bottomrule
\end{tabular}
}
\caption{Pearson Correlation Coefficients and \(p\)-values}
\end{table}
We sampled Mean Squared Error (MSE), Focus Ratio (FR), and Memory Entropy (ME) at various epochs during multiple training iterations, resulting in a dataset comprising 300 sample points. Based on this data, we conducted statistical significance tests for FR and ME, as well as analyzed their correlations with MSE.

The analysis reveals significant negative correlations between Mean Squared Error (MSE) and both Focus Ratio (FR) and Memory Entropy (ME). Specifically, the Pearson correlation coefficient between MSE and FR is \(r = -0.5096\) (\(p < 0.001\)), indicating a moderate inverse relationship. Similarly, the correlation between MSE and ME is \(r = -0.8000\) (\(p < 0.001\)), demonstrating a robust negative association. These findings suggest that higher Focus Ratio and lower Memory Entropy are closely linked to reduced prediction errors, highlighting the critical role of attentional focus and efficient memory processes in enhancing predictive performance. The results underscore the importance of optimizing cognitive mechanisms, such as improving attention allocation and refining memory structures, to achieve greater accuracy in predictive tasks.

\subsection{Detail Comparison of Improvements}

To demonstrate that pre-training Mamba blocks with RCL can effectively enhance the temporal prediction capabilities of Mamba-based models, we present the performance improvements of four Mamba-based models after using pre-trained parameters. We conducted extensive tesWting on six datasets, each with an input length of 96 and prediction lengths of \( \{96, 192, 336, 720\} \). To clearly illustrate the performance improvements, we provide the percentage increase in MSE and MAE when using pre-trained parameters compared to not using them, as shown by the up-rate in Table \ref{tab:dcomp}. 

The results indicate that, for the vast majority of datasets and prediction lengths, the parameters obtained through our method enhance the predictive performance of Mamba-based models,  demonstrating that our approach is generally effective. By pre-training a Mamba block and using the pre-trained parameters to initialize all mamba blocks in Mamba-based model, the original model's temporal prediction performance can be significantly improved.

\subsection{Standard Deviation}\label{std}
We provided the standard deviation of experimental results across multiple datasets. As Shown in Table \ref{tab:variance}.It can be observed that using RCL parameters for initialization results in a standard deviation similar to not using them, indicating that our method enhances performance without introducing additional instability.

\begin{table*}[htbp]
  \centering
    \resizebox{0.9\linewidth}{!}{
   
    \begin{tabular}{c|c|cc|cc|cc|cc}
    \toprule
    \multicolumn{2}{c|}{Model} & \multicolumn{2}{c|}{\textbf{TimeMachine*}} & \multicolumn{2}{c|}{TimeMachine} & \multicolumn{2}{c|}{\textbf{Bi-Mamba*}} & \multicolumn{2}{c}{Bi-Mamba} \\
    \midrule
    \multicolumn{2}{c|}{Metric} & MAE   & MSE   & MAE   & MSE   & MAE   & MSE   & MAE   & MSE \\
    \midrule
    \multirow{4}[2]{*}{ETTh1} & 96    & ±0.001 & ±0.002 & ±0.002 & ±0.002 & ±0.003 & ±0.007 & ±0.005 & ±0.007 \\
          & 192   & ±0.002 & ±0.005 & ±0.004 & ±0.008 & ±0.003 & ±0.006 & ±0.006 & ±0.007 \\
          & 336   & ±0.002 & ±0.006 & ±0.003 & ±0.005 & ±0.006 & ±0.005 & ±0.009 & ±0.013 \\
          & 720   & ±0.006 & ±0.006 & ±0.009 & ±0.013 & ±0.007 & ±0.009 & ±0.008 & ±0.015 \\
    \midrule
    \multirow{4}[1]{*}{ETTh2} & 96    & ±0.001 & ±0.001 & ±0.005 & ±0.007 & ±0.001 & ±0.002 & ±0.001 & ±0.001 \\
          & 192   & ±0.002 & ±0.001 & ±0.006 & ±0.006 & ±0.002 & ±0.002 & ±0.001 & ±0.002 \\
          & 336   & ±0.004 & ±0.006 & ±0.009 & ±0.013 & ±0.003 & ±0.003 & ±0.003 & ±0.004 \\
          & 720   & ±0.006 & ±0.006 & ±0.010 & ±0.009 & ±0.006 & ±0.006 & ±0.007 & ±0.010 \\
    \midrule
    \multirow{4}[0]{*}{ETTm1} & 96    & ±0.002 & ±0.002 & ±0.002 & ±0.001 & ±0.002 & ±0.002 & ±0.003 & ±0.002 \\
          & 192   & ±0.003 & ±0.006 & ±0.003 & ±0.003 & ±0.002 & ±0.004 & ±0.003 & ±0.004 \\
          & 336   & ±0.003 & ±0.005 & ±0.004 & ±0.004 & ±0.003 & ±0.005 & ±0.004 & ±0.005 \\
          & 720   & ±0.006 & ±0.009 & ±0.005 & ±0.007 & ±0.004 & ±0.004 & ±0.003 & ±0.004 \\
    \midrule
    \multirow{4}[1]{*}{ETTm2} & 96    & ±0.001 & ±0.001 & ±0.001 & ±0.001 & ±0.001 & ±0.001 & ±0.001 & ±0.001 \\
          & 192   & ±0.001 & ±0.002 & ±0.001 & ±0.001 & ±0.001 & ±0.001 & ±0.001 & ±0.001 \\
          & 336   & ±0.003 & ±0.005 & ±0.002 & ±0.006 & ±0.002 & ±0.003 & ±0.002 & ±0.002 \\
          & 720   & ±0.005 & ±0.005 & ±0.004 & ±0.006 & ±0.004 & ±0.005 & ±0.003 & ±0.004 \\
    \midrule
    \multirow{4}[1]{*}{Traffic} & 96    & ±0.005 & ±0.004 & ±0.004 & ±0.003 & ±0.002 & ±0.002 & ±0.003 & ±0.005 \\
          & 192   & ±0.006 & ±0.005 & ±0.003 & ±0.003 & ±0.001 & ±0.002 & ±0.001 & ±0.002 \\
          & 336   & ±0.006 & ±0.006 & ±0.005 & ±0.005 & ±0.002 & ±0.002 & ±0.001 & ±0.003 \\
          & 720   & ±0.005 & ±0.007 & ±0.006 & ±0.009 & ±0.003 & ±0.003 & ±0.001 & ±0.002 \\
    \midrule
    \multirow{4}[1]{*}{Electricity} & 96    & ±0.001 & ±0.002 & ±0.001 & ±0.002 & ±0.001 & ±0.001 & ±0.001 & ±0.001 \\
          & 192   & ±0.001 & ±0.001 & ±0.002 & ±0.002 & ±0.001 & ±0.001 & ±0.002 & ±0.002 \\
          & 336   & ±0.002 & ±0.002 & ±0.001 & ±0.001 & ±0.001 & ±0.002 & ±0.001 & ±0.003 \\
          & 720   & ±0.001 & ±0.002 & ±0.001 & ±0.001 & ±0.001 & ±0.002 & ±0.002 & ±0.002 \\
    \bottomrule
    \end{tabular}%
      }
    \caption{Standard deviation of experimental results}
  \label{tab:variance}%
\end{table*}%

\subsection{Generalization Experiments on Transformer and RNN}




To evaluate the efficacy of our Recurrent Contrastive Learning (RCL) framework and associated training methodologies, we conducted experiments on both a single-layer Transformer model and a Recurrent Neural Network (RNN) model. During the pretraining phase, we maintained identical paTrameter configurations as those employed in the primary experiments of this study. Specifically, Gaussian noise was applied, and three iterations of augmentation were performed for each timestamp token to enhance robustness. The models were pretrained until convergence, after which the pretrained parameters were transferred to initialize the downstream models for time series prediction tasks. To ensure experimental consistency, all other settings for the downstream tasks were aligned with those of the main experiment, facilitating equitable comparisons across datasets.

The results of these experiments are systematically presented in Table \ref{tab:comp_freeze} and \ref{tab:comp_model}. Table \ref{tab:comp_freeze} investigates the adaptability of the pretrained parameters when integrated into the downstream models. Meanwhile, Table \ref{tab:comp_model} quantifies the performance improvements, reporting the percentage changes in Mean Absolute Error (MAE) and Mean Squared Error (MSE) metrics for both models, with and without RCL pretraining.

Analysis of Table \ref{tab:comp_model} reveals that RCL pretraining consistently enhances the performance of RNN models across all evaluated datasets, with negligible adverse effects. Specifically, RCL yields an average improvement exceeding 7\% in both MAE and MSE metrics on the prediction set, underscoring its substantial positive impact on RNN-based architectures for spatiotemporal prediction tasks. However, this enhancement may be constrained by inherent limitations of the RNN model, notably its limited memory capacity compared to advanced architectures like the Mamba model. Furthermore, RNNs exhibit reduced sensitivity to long time series after three augmentation iterations, resulting in relatively modest overall improvements.

In contrast, the Transformer model demonstrates a more limited response to RCL pretraining, with an average improvement in MAE and MSE metrics of less than 3\%. Notably, much of the observed improvement is attributable to a single dataset, suggesting potential overfitting. The influence of downstream training on Transformer performance appears minimal, whereas the initialization parameters exert a significant effect. Simultaneously, RCL repeats each token in sequences, tripling the input length, which consequently increases the GPU memory cost by approximately three times during the pretraining stage compared to RNNs. This substantial increase in computational resources limits the practical applicability of RCL pretraining for Transformer models. Given the inherent randomness in experimental outcomes, the modest overall improvement does not provide sufficient evidence to confirm a robust performance enhancement, indicating that RCL pretraining may not effectively optimize Transformer models for time series prediction tasks.

These findings align with the hypotheses posited in the main paper. We propose an intuitive interpretation: RCL enhances model selectivity by leveraging repeated time steps and constructing positive and negative sample pairs through token-level contrastive learning. This mechanism is particularly advantageous for models that process sequences linearly, such as RNNs and the Mamba architecture, which benefit from the structured augmentation of temporal data. Conversely, the Transformer’s self-attention mechanism, characterized by its $O(n^2)$ complexity, enables comprehensive focus across the entire sequence, rendering it less dependent on repeated time steps. Consequently, the Transformer is less likely to derive significant benefits from RCL pretraining, which may explain the observed variability in its experimental outcomes.

\begin{table*}[htbp]
  \centering

  \resizebox{\linewidth}{!}{
    \begin{tabular}{c|c|c|cc|cc|cc|cc|cc|cc}
    \toprule
    \multicolumn{3}{r|}{\multirow{2}[4]{*}{}} & \multicolumn{2}{c|}{ETTh1} & \multicolumn{2}{c|}{ETTh2} & \multicolumn{2}{c|}{ETTm1} & \multicolumn{2}{c|}{ETTm2} & \multicolumn{2}{c|}{Traffic} & \multicolumn{2}{c}{Electricity} \\
\cmidrule{4-15}    \multicolumn{3}{r|}{} & MAE   & MSE   & MAE   & MSE   & MAE   & MSE   & MAE   & MSE   & MAE   & MSE   & MAE   & MSE \\
    \midrule
    \multirow{12}[8]{*}{Mamba } & \multirow{3}[2]{*}{96}  & w/o   & 0.6546  & 0.7672  & 1.4013  & 2.8442  & 0.5053  & 0.5432  & 0.5763  & 0.6008  & 0.4939  & 1.0279  & 0.4232  & 0.3926  \\
          &       & w     & 0.5974  & 0.6542  & 1.1536  & 2.0506  & 0.4798  & 0.4946  & 0.5646  & 0.5677  & 0.4604  & 0.9076  & 0.4168  & 0.3879  \\
          &       & up-rate\% & \textbf{8.7382 } & \textbf{14.729 } & \textbf{17.676 } & \textbf{27.902 } & \textbf{5.0465 } & \textbf{8.9470 } & \textbf{2.0302 } & \textbf{5.5093 } & \textbf{6.7827 } & \textbf{11.704 } & \textbf{1.5123 } & \textbf{1.1.971 } \\
\cmidrule{2-15}          & \multirow{3}[2]{*}{192} & w/o   & 0.6298  & 0.7115  & 1.2371  & 2.1642  & 0.5126  & 0.5866  & 0.6670  & 0.8471  & 0.5617  & 1.1962  & 0.4298  & 0.4053  \\
          &       & w     & 0.6021  & 0.7127  & 1.0509  & 1.9490  & 0.4970  & 0.5524  & 0.5655  & 0.5573  & 0.5610  & 1.1877  & 0.4288  & 0.4130  \\
          &       & up-rate\% & 4.3982  & -0.1687  & \textbf{15.0513 } & \textbf{9.9436 } & \textbf{3.0433 } & \textbf{5.8302 } & \textbf{15.2174 } & \textbf{34.2108 } & \textbf{0.1246 } & \textbf{0.7106 } & 0.2327  & -1.8998  \\
\cmidrule{2-15}          & \multirow{3}[2]{*}{336} & w/o   & 0.6383  & 0.7210  & 1.2341  & 2.1528  & 0.8172  & 1.4569  & 0.7052  & 0.9220  & 0.6025  & 1.3079  & 0.4354  & 0.4108  \\
          &       & w     & 0.6084  & 0.7145  & 1.0497  & 1.9485  & 0.8008  & 1.4479  & 0.6270  & 0.6842  & 0.5848  & 1.2560  & 0.4324  & 0.4176  \\
          &       & up-rate\% & \textbf{4.6843 } & \textbf{0.9015 } & \textbf{14.9421 } & \textbf{9.4900 } & \textbf{2.0069 } & \textbf{0.6178 } & \textbf{11.0891 } & \textbf{25.7918 } & \textbf{2.9378 } & \textbf{3.9682 } & 0.6890  & -1.6553  \\
\cmidrule{2-15}          & \multirow{3}[2]{*}{720} & w/o   & 0.6776  & 0.7727  & 1.2206  & 2.1005  & 0.8235  & 1.4557  & 0.7374  & 0.9942  & 0.4893  & 1.0108  & 0.4529  & 0.4326  \\
          &       & w     & 0.6461  & 0.7556  & 1.0541  & 1.9537  & 0.8142  & 1.4588  & 0.6682  & 0.7811  & 0.4645  & 0.9189  & 0.4447  & 0.4320  \\
          &       & up-rate\% & \textbf{4.6488 } & \textbf{2.2130 } & \textbf{13.6408 } & \textbf{6.9888 } & 1.1293  & -0.2130  & \textbf{9.3843 } & \textbf{21.4343 } & \textbf{5.0685 } & \textbf{9.0918 } & \textbf{1.8106 } & \textbf{0.1387 } \\
    \midrule
    \multirow{12}[8]{*}{iMamba } & \multirow{3}[2]{*}{96} & w/o   & 0.4987  & 0.4928  & 0.6926  & 0.9084  & 0.4316  & 0.3998  & 0.4160  & 0.3666  & 0.3234  & 0.6538  & 0.2627  & 0.1857  \\
          &       & w     & 0.4472  & 0.4278  & 0.6833  & 0.8595  & 0.3970  & 0.3669  & 0.3304  & 0.2469  & 0.2913  & 0.6003  & 0.2597  & 0.1827  \\
          &       & up-rate\% & \textbf{10.3268 } & \textbf{13.1899 } & \textbf{1.3428 } & \textbf{5.3831 } & \textbf{8.0167 } & \textbf{8.2291 } & \textbf{20.5769 } & \textbf{32.6514 } & \textbf{9.9258 } & \textbf{8.1829 } & \textbf{1.1420 } & \textbf{1.6155 } \\
\cmidrule{2-15}          & \multirow{3}[2]{*}{192} & w/o   & 0.5075  & 0.5320  & 1.0228  & 1.8207  & 0.4500  & 0.4390  & 0.4973  & 0.4949  & 0.3129  & 0.6354  & 0.2801  & 0.2047  \\
          &       & w     & 0.4871  & 0.5143  & 0.9430  & 1.5825  & 0.4356  & 0.4174  & 0.4763  & 0.4557  & 0.3091  & 0.6335  & \multicolumn{1}{r}{0.2788 } & \multicolumn{1}{r}{0.2025 } \\
          &       & up-rate\% & \textbf{4.0197 } & \textbf{3.3271 } & \textbf{7.8021 } & \textbf{13.0829 } & \textbf{3.2000 } & \textbf{4.9203 } & \textbf{4.2228 } & \textbf{7.9208 } & \textbf{1.2144 } & \textbf{0.2990 } & \textbf{0.4641 } & \textbf{1.0747 } \\
\cmidrule{2-15}          & \multirow{3}[2]{*}{336} & w/o   & 0.5125  & 0.5498  & 1.0727  & 2.0417  & 0.4909  & 0.5085  & 0.7932  & 1.0322  & 0.3233  & 0.6605  & 0.2987  & 0.2238  \\
          &       & w     & 0.4750  & 0.4992  & 0.9913  & 1.7052  & 0.4677  & 0.4998  & 0.5854  & 0.6272  & 0.3216  & 0.6645  & 0.2975  & 0.2222  \\
          &       & up-rate\% & \textbf{7.3171 } & \textbf{9.2033 } & \textbf{7.5883 } & \textbf{16.4814 } & \textbf{4.7260 } & \textbf{1.7109 } & \textbf{26.1977 } & \textbf{39.2366 } & 0.5258  & -0.6056  & \textbf{0.4017 } & \textbf{0.7149 } \\
\cmidrule{2-15}          & \multirow{3}[2]{*}{720} & w/o   & 0.5418  & 0.5818  & 1.0534  & 1.8199  & 0.6238  & 0.7306  & 1.0698  & 2.0298  & 0.3486  & 0.7105  & \multicolumn{1}{r}{0.3342 } & \multicolumn{1}{r}{0.2683 } \\
          &       & w     & 0.5391  & 0.5640  & 1.0172  & 1.7220  & 0.5120  & 0.5534  & 0.9936  & 1.5644  & 0.3475  & 0.7172  & \multicolumn{1}{r}{0.3323 } & \multicolumn{1}{r}{0.2627 } \\
          &       & up-rate\% & \textbf{0.4983 } & \textbf{3.0595 } & \textbf{3.4365 } & \textbf{5.3794 } & \textbf{17.9224 } & \textbf{24.2540 } & \textbf{7.1228 } & \textbf{22.9284 } & 0.3155  & -0.9430  & \textbf{0.5685 } & \textbf{2.0872 } \\
    \midrule
    \multirow{12}[8]{*}{TimeMachine} & \multirow{3}[2]{*}{96} & w/o   & 0.3905  & 0.3833  & 0.3344  & 0.2911  & 0.3606  & 0.3342  & 0.2525  & 0.1746  & 0.3064  & 0.4983  & 0.2611  & 0.1872  \\
          &       & w     & 0.3869  & 0.3787  & 0.3298  & 0.2822  & 0.3458  & 0.3179  & 0.2508  & 0.1731  & 0.2991  & 0.4844  & 0.2586  & 0.1826  \\
          &       & up-rate\% & \textbf{0.9219 } & \textbf{1.2001 } & \textbf{1.3756 } & \textbf{3.0574 } & \textbf{4.1043 } & \textbf{4.8773 } & \textbf{0.6733 } & \textbf{0.8591 } & \textbf{2.3825 } & \textbf{2.7895 } & \textbf{0.9575 } & \textbf{2.4573 } \\
\cmidrule{2-15}          & \multirow{3}[2]{*}{192} & w/o   & 0.4225  & 0.4401  & 0.3851  & 0.3685  & 0.3785  & 0.3787  & 0.2941  & 0.2381  & 0.2740  & 0.4170  & 0.2500  & 0.1580  \\
          &       & w     & 0.4202  & 0.4399  & 0.3821  & 0.3551  & 0.3770  & 0.3750  & 0.2930  & 0.2381  & 0.2732  & 0.4115  & 0.2460  & 0.1520  \\
          &       & up-rate\% & \textbf{0.5444 } & \textbf{0.0454 } & \textbf{0.7790 } & \textbf{3.6364 } & \textbf{0.3963 } & \textbf{0.9770 } & \textbf{0.3740 } & \textbf{0.0000 } & \textbf{0.2920 } & \textbf{1.3189 } & \textbf{1.6000 } & \textbf{3.7975 } \\
\cmidrule{2-15}          & \multirow{3}[2]{*}{336} & w/o   & 0.4458  & 0.4902  & 0.4281  & 0.4206  & 0.3937  & 0.4010  & 0.3371  & 0.3066  & 0.2810  & 0.4330  & 0.2680  & 0.1720  \\
          &       & w     & 0.4419  & 0.4824  & 0.4201  & 0.4119  & 0.3867  & 0.3956  & 0.3327  & 0.2991  & 0.2790  & 0.4290  & 0.2610  & 0.1690  \\
          &       & up-rate\% & \textbf{0.8748 } & \textbf{1.5912 } & \textbf{1.8687 } & \textbf{2.0685 } & \textbf{1.7780 } & \textbf{1.3466 } & \textbf{1.3053 } & \textbf{2.4462 } & \textbf{0.7117 } & \textbf{0.9238 } & \textbf{2.6119 } & \textbf{1.7442 } \\
\cmidrule{2-15}          & \multirow{3}[2]{*}{720} & w/o   & 0.4702  & 0.4959  & 0.4386  & 0.4243  & 0.4310  & 0.4670  & 0.3940  & 0.4073  & 0.3000  & 0.4670  & 0.2980  & 0.2070  \\
          &       & w     & 0.4656  & 0.4883  & 0.4295  & 0.4119  & 0.4291  & 0.4552  & 0.3920  & 0.4018  & 0.2980  & 0.4590  & 0.2950  & 0.2010  \\
          &       & up-rate\% & \textbf{0.9783 } & \textbf{1.5326 } & \textbf{2.0748 } & \textbf{2.9225 } & \textbf{0.4408 } & \textbf{2.5268 } & \textbf{0.5076 } & \textbf{1.3504 } & \textbf{0.6667 } & \textbf{1.7131 } & \textbf{1.0067 } & \textbf{2.8986 } \\
    \midrule
    \multirow{12}[8]{*}{Bi-Mamba} & \multirow{3}[2]{*}{96} & w/o   & 0.3948  & 0.3813  & 0.3443 & 0.2937  & 0.3641  & 0.3319  &   0.2704    &   0.1883    &   0.2786    &   0.587    &   0.2629    & 0.185 \\
          &       & w     & 0.3893  & 0.3794  & 0.3462  & 0.2955  & 0.3578 & 0.3316  &   0.2707    &   0.1857    &   0.2761    &   0.5787    &   0.2611    & 0.1818 \\
          &       & up-rate\% & \textbf{1.3931 } & \textbf{0.4983 } & \textbf{1.7303 } & \textbf{0.0904 } & \textbf{0.9829 } & \textbf{1.2814 } & -0.1109  & \textbf{1.3808 } & \textbf{0.8973 } & \textbf{1.4140 } & \textbf{0.6847 } & \textbf{1.7280 } \\
\cmidrule{2-15}          & \multirow{3}[2]{*}{192} & w/o   & 0.4280  & 0.4270  & 0.3977  & 0.3772  & 0.3894  & 0.3780  & 0.3145  & 0.2572  & 0.3057  & 0.6301  & 0.2715  & 0.1914  \\
          &       & w     & 0.4210  & 0.4250  & 0.3935  & 0.3733  & 0.3840  & 0.3692  & 0.3131  & 0.2544  & 0.3081  & 0.6250  & 0.2698  & 0.1881  \\
          &       & up-rate\% & \textbf{1.6355 } & \textbf{0.4684 } & \textbf{1.0561 } & \textbf{1.0339 } & \textbf{1.3867 } & \textbf{2.3280 } & \textbf{0.4452 } & \textbf{1.0886 } & -0.7851  & 0.8094  & \textbf{0.6262 } & \textbf{1.7241 } \\
\cmidrule{2-15}          & \multirow{3}[2]{*}{336} & w/o   & 0.4593  & 0.4838  & 0.4340  & 0.4354  & 0.4119  & 0.4045  & 0.3871  & 0.3915  & 0.3068  & 0.6585  & 0.2896  & 0.2117  \\
          &       & w     & 0.4563  & 0.4805  & 0.4286  & 0.4344  & 0.4069  & 0.4036  & 0.3644  & 0.3158  & 0.3107  & 0.6659  & 0.2831  & 0.1999  \\
          &       & up-rate\% & \textbf{0.6532 } & \textbf{0.6821 } & \textbf{1.2442 } & \textbf{0.2297 } & \textbf{1.2139 } & \textbf{0.2225 } & \textbf{5.8641 } & \textbf{19.3359 } & -1.2712  & -1.1238  & \textbf{2.2445 } & \textbf{5.5739 } \\
\cmidrule{2-15}          & \multirow{3}[2]{*}{720} & w/o   & 0.4963  & 0.5164  & 0.5970  & 0.7150  & 0.4517  & 0.4659  & 0.4300  & 0.4292  & 0.3384  & 0.7015  & 0.3228  & 0.2591  \\
          &       & w     & 0.4960  & 0.4962  & 0.6020  & 0.7310  & 0.4413  & 0.4579  & 0.4131  & 0.4044  & 0.3364  & 0.6894  & 0.3174  & 0.2547  \\
          &       & up-rate\% & \textbf{0.0604 } & \textbf{3.9117 } & -0.8375  & -2.2378  & \textbf{2.3024 } & \textbf{1.7171 } & \textbf{3.9302 } & \textbf{5.7782 } & \textbf{0.5910 } & \textbf{1.7249 } & \textbf{1.6729 } & \textbf{1.6982 } \\
    \bottomrule
    \end{tabular}%
}
  \caption{Detail Comparison of performance improvement by replacing parameters obtained by RCL. w/o denotes no parameter replacement, w denotes parameter replacement, and up-rate represents the improvement rate.}
  \label{tab:dcomp}%
\end{table*}%

\begin{table}[htbp]
  \centering
  
  \resizebox{\linewidth}{!}{

        \begin{tabular}{c|c|cc|cc|cc|cc|cc|cc}
    \toprule
    \multicolumn{1}{c}{\multirow{2}[4]{*}{}} & \multirow{2}[4]{*}{} & \multicolumn{2}{c|}{ETTh1} & \multicolumn{2}{c|}{ETTh2} & \multicolumn{2}{c|}{ETTm1} & \multicolumn{2}{c|}{ETTm2} & \multicolumn{2}{c|}{Electricity} & \multicolumn{2}{c}{Traffic} \\
\cmidrule{3-14}    \multicolumn{1}{c}{} &       & MAE   & MSE   & MAE   & MSE   & MAE   & MSE   & MAE   & MSE   & MAE   & MSE   & MAE   & MSE \\
    \midrule
    \multirow{3}[2]{*}{Transformer W ALL} & 96    & 0.5094  & 0.5614  & \textbf{0.7226 } & 0.8833  & 0.7068  & 0.8778  & 0.4329  & 0.3576  & 0.4038  & 0.3481  & 0.4146  & 0.8386  \\
          & 192   & 0.6462  & \textbf{0.7544 } & \textbf{1.0953 } & \textbf{1.7122 } & 0.7423  & 0.9701  & 0.7903  & 1.0136  & 0.4677  & 0.4527  & 0.5268  & \textbf{1.1109 } \\
          & 336   & 0.7354  & 0.9376  & \textbf{1.4186 } & \textbf{3.1809 } & \textbf{1.1192 } & \textbf{1.9197 } & 1.2812  & 2.4169  & 0.4417  & 0.4383  & 0.5335  & 1.1198  \\
    \midrule
    \multirow{3}[2]{*}{Transformer W None} & 96    & 0.5126  & \textbf{0.5465 } & 0.9167  & 1.5182  & 0.7091  & 0.8679  & \textbf{0.3689 } & 0.2845  & \textbf{0.3812 } & \textbf{0.3126 } & 0.4097  & 0.8439  \\
          & 192   & \textbf{0.6176 } & 0.8058  & 1.2510  & 2.6254  & 0.7367  & 0.9480  & 0.5980  & 0.6023  & 0.4672  & 0.4629  & 0.5382  & 1.1644 \\
          & 336   & 0.8519  & 1.1444  & 1.6312  & 3.6907  & 1.2601  & 2.4396  & 1.1218  & 2.1189  & 0.4448  & \textbf{0.4377 } & 0.5371  & 1.1297  \\
    \midrule
    \multirow{3}[2]{*}{Transformer} & 96    & \textbf{0.4852 } & 0.5533  & 0.7365  & \textbf{0.8280 } & \textbf{0.6377 } & \textbf{0.7231 } & 0.3684  & \textbf{0.2781 } & 0.4548  & 0.4180  & \textbf{0.4060} & \textbf{0.8270} \\
          & 192   & 0.6350  & 0.8073  & 1.6950  & 3.7041  & \textbf{0.6928 } & \textbf{0.8255 } & \textbf{0.5105 } & \textbf{0.4832 } & \textbf{0.4493 } & \textbf{0.4197 } & \textbf{0.5221} & 1.1296 \\
          & 336   & \textbf{0.7271 } & \textbf{0.8731 } & 1.7984  & 4.4986  & 1.2989  & 2.6087  & \textbf{1.0137 } & \textbf{1.6190 } & \textbf{0.4473 } & 0.4430  & \textbf{0.5315} & \textbf{1.1158} \\
    \midrule
    \multirow{3}[2]{*}{RNN W ALL} & 96    & 0.6267 & 0.7201 & \textbf{1.0847} & 2.1172 & 0.7138 & 0.8417 & \textbf{0.4984} & \textbf{0.4353} & 0.4087 & 0.3642 & \textbf{0.4009} & \textbf{0.8454} \\
          & 192   & \textbf{0.7001} & \textbf{0.8089} & \textbf{1.6082} & \textbf{3.4259} & 0.7072 & 0.8382 & \textbf{0.7737} & \textbf{1.0554} & \textbf{0.4185} & \textbf{0.3972} & \textbf{0.5248} & \textbf{1.1147} \\
          & 336   & 0.7370 & \textbf{0.8754} & \textbf{1.3153} & \textbf{2.6742} & 1.0757 & 1.7429 & \textbf{1.1009} & \textbf{1.9765} & \textbf{0.4339} & \textbf{0.4224} & \textbf{0.5414} & \textbf{1.1529} \\
    \midrule
    \multirow{3}[2]{*}{RNN W None} & 96    & 0.6053 & 0.6806 & 1.1511 & 2.1204 & \textbf{0.5866} & \textbf{0.6364} & 0.5999 & 0.7086 & \textbf{0.3899} & \textbf{0.3310} & 0.4241 & 0.9131 \\
          & 192   & 0.7356 & 0.8716 & 2.1479 & 5.8670 & \textbf{0.6686} & \textbf{0.7873} & 0.9108 & 1.3498 & 0.4426 & 0.4222 & 0.5211 & 1.1184 \\
          & 336   & 0.7894 & 0.9665 & 1.5554 & 3.1801 & \textbf{0.9510} & \textbf{1.4781} & 1.1995 & 2.4811 & 0.4513 & 0.4534 & 0.5618 & 1.2234 \\
    \midrule
    \multirow{3}[2]{*}{RNN} & 96    & \textbf{0.5746} & \textbf{0.6558} & 1.1737 & \textbf{2.1077} & 0.6333 & 0.7266 & 0.5409 & 0.5519 & 0.4398 & 0.4035 & 0.4111 & 0.8733 \\
          & 192   & 0.7273 & 0.8619 & 1.7691 & 4.1801 & 0.7253 & 0.8969 & 0.8243 & 1.0677 & 0.4248 & 0.4045 & 0.5275 & 1.1347 \\
          & 336   & \textbf{0.7362} & 0.9411 & 1.7274 & 3.9867 & 1.0066 & 1.5956 & 1.3114 & 2.8928 & 0.4493 & 0.4472 & 0.5707 & 1.2280 \\
    \bottomrule
    \end{tabular}%
    }
  \caption{Detail Comparison of experiments between initializing models parameters from RCL pretraining, and whether to fix the initlized parameters in downstream tasks. w/o denotes no parameter replacement, All denotes all loaded parameter is fixed, and None represents the all loaded parameters can be modified in downstream tasks}
  \label{tab:comp_freeze}%
\end{table}%

\begin{table}[htbp]
  \centering
  
 \resizebox{\linewidth}{!}{
    \begin{tabular}{c|r|l|c|c|c|c|c|c|c|c|c|c|c|c|}
    \toprule
    \multicolumn{3}{r|}{\multirow{2}[3]{*}{}} & \multicolumn{2}{c|}{ETTh1} & \multicolumn{2}{c|}{ETTh2} & \multicolumn{2}{c|}{ETTm1} & \multicolumn{2}{c|}{ETTm2} & \multicolumn{2}{c|}{Electricity} & \multicolumn{2}{c|}{Traffic} \\
\cmidrule{4-15}    \multicolumn{3}{r|}{} & MAE   & MSE   & MAE   & MSE   & MAE   & MSE   & MAE   & MSE   & MAE   & MSE   & MAE   & MSE \\
    \midrule
    \multirow{9}[5]{*}{Transformer} & \multirow{3}[1]{*}{96} & w     & 0.5094  & 0.5465  & 0.7226  & 0.8833  & 0.7068  & 0.8679  & 0.3689  & 0.2845  & 0.3812  & 0.3126  & 0.4097  & 0.8386  \\
          &       & w/o   & 0.4852  & 0.5533  & 0.7365  & 0.8280  & 0.6377  & 0.7231  & 0.3684  & 0.2781  & 0.4548  & 0.4180  & 0.4060  & 0.8270  \\
          &       & up-rate\% & -5.0025 & \textbf{1.2398 } & \textbf{1.8912 } & -6.6805 & -10.8314 & -20.0306 & -0.1282 & -2.3160 & \textbf{16.1845 } & \textbf{25.2042 } & -0.9150 & -1.4060 \\
\cmidrule{2-15}          & \multirow{3}[2]{*}{192} & w     & 0.6176  & 0.7544  & 1.0953  & 1.7122  & 0.7367  & 0.9480  & 0.5980  & 0.6023  & 0.4672  & 0.4527  & 0.5268  & 1.1109  \\
          &       & w/o   & 0.6350  & 0.8073  & 1.6950  & 3.7041  & 0.6928  & 0.8255  & 0.5105  & 0.4832  & 0.4493  & 0.4197  & 0.5221  & 1.1296  \\
          &       & up-rate\% & \textbf{2.7354 } & \textbf{6.5577 } & \textbf{35.3795 } & \textbf{53.7760 } & -6.3294 & -14.8469 & -17.1455 & -24.6520 & -3.9822 & -7.8734 & -0.8892 & \textbf{1.6616 } \\
\cmidrule{2-15}          & \multirow{3}[2]{*}{336} & w     & 0.7354  & 0.9376  & 1.4186  & 3.1809  & 1.1192  & 1.9197  & 1.1218  & 2.1189  & 0.4417  & 0.4377  & 0.5335  & 1.1198  \\
          &       & w/o   & 0.7271  & 0.8731  & 1.7984  & 4.4986  & 1.2989  & 2.6087  & 1.0137  & 1.6190  & 0.4473  & 0.4430  & 0.5315  & 1.1158  \\
          &       & up-rate\% & -1.1479 & -7.3808 & \textbf{21.1174 } & \textbf{29.2910 } & \textbf{13.8329 } & \textbf{26.4096 } & -10.6680 & -30.8726 & \textbf{1.2409 } & \textbf{1.2117 } & -0.3655 & -0.3641 \\
    \midrule
    \multirow{9}[6]{*}{RNN} & \multirow{3}[2]{*}{96} & w     & 0.6053  & 0.6806  & 1.0847  & 2.1172  & 0.5866  & 0.6364  & 0.4984  & 0.4353  & 0.3899  & 0.3310  & 0.4009  & 0.8454  \\
          &       & w/o   & 0.5746  & 0.6558  & 1.1737  & 2.1077  & 0.6333  & 0.7266  & 0.5409  & 0.5519  & 0.4398  & 0.4035  & 0.4111  & 0.8733  \\
          &       & up-rate\% & -5.3462 & -3.7867 & \textbf{7.5868 } & -0.4468 & \textbf{7.3619 } & \textbf{12.4108 } & \textbf{7.8511 } & \textbf{21.1200 } & \textbf{11.3497 } & \textbf{17.9789 } & \textbf{2.4792 } & \textbf{3.1968 } \\
\cmidrule{2-15}          & \multirow{3}[2]{*}{192} & w     & 0.7001  & 0.8089  & 1.6082  & 3.4259  & 0.6686  & 0.7873  & 0.7737  & 1.0554  & 0.4185  & 0.3972  & 0.5211  & 1.1147  \\
          &       & w/o   & 0.7273  & 0.8619  & 1.7691  & 4.1801  & 0.7253  & 0.8969  & 0.8243  & 1.0677  & 0.4248  & 0.4045  & 0.5275  & 1.1347  \\
          &       & up-rate\% & \textbf{3.7365 } & \textbf{6.1514 } & \textbf{9.0946 } & \textbf{18.0427 } & \textbf{7.8229 } & \textbf{12.2164 } & \textbf{6.1402 } & \textbf{1.1543 } & \textbf{1.4861 } & \textbf{1.8081 } & \textbf{1.2209 } & \textbf{1.7621 } \\
\cmidrule{2-15}          & \multirow{3}[2]{*}{336} & w     & 0.7370  & 0.8754  & 1.3153  & 2.6742  & 0.9510  & 1.4781  & 1.1009  & 1.9765  & 0.4339  & 0.4224  & 0.5414  & 1.1529  \\
          &       & w/o   & 0.7362  & 0.9411  & 1.7274  & 3.9867  & 1.0066  & 1.5956  & 1.3114  & 2.8928  & 0.4493  & 0.4472  & 0.5707  & 1.2280  \\
          &       & up-rate\% & -0.1038 & \textbf{6.9792 } & \textbf{23.8573 } & \textbf{32.9220 } & \textbf{5.5162 } & \textbf{7.3632 } & \textbf{16.0520 } & \textbf{31.6742 } & \textbf{3.4215 } & \textbf{5.5484 } & \textbf{5.1283 } & \textbf{6.1132 } \\
    \bottomrule
    \end{tabular}%
   }
   \caption{Detail Comparison of performance improvement by replacing parameters obtained by RCL on Transformer and RNN models. w/o denotes no parameter replacement, w denotes parameter replacement, and up-rate represents the improvement rate}
  \label{tab:comp_model}%
\end{table}%

\end{document}